# Stage 4 validation of the Satellite Image Automatic Mapper lightweight computer program for Earth observation Level 2 product generation – Part 1: Theory

A. Baraldi[a,c,*], M. L. Humber[b], D. Tiede[c] and S. Lang[c]

[a] *Department of Agricultural and Food Sciences, University of Naples Federico II, Portici (NA), Italy.*
[b] *Department of Geographical Sciences, University of Maryland, College Park, MD 20742, USA.*
[c] *Department of Geoinformatics – Z_GIS, University of Salzburg, Salzburg 5020, Austria.*

*Corresponding author. Email: andrea6311@gmail.com

*Abstract* - The European Space Agency (ESA) defines an Earth Observation (EO) Level 2 product as a multi-spectral (MS) image corrected for geometric, atmospheric, adjacency and topographic effects, stacked with its data-derived scene classification map (SCM), whose legend includes quality layers such as cloud and cloud-shadow. No ESA EO Level 2 product has ever been systematically generated at the ground segment. To contribute toward filling an analytic and pragmatic information gap from EO big data to the ESA EO Level 2 product, an original Stage 4 validation (*Val*) of the Satellite Image Automatic Mapper (SIAM) lightweight computer program was conducted by independent means on an annual Web-Enabled Landsat Data (WELD) image composite time-series of the conterminous U.S. The non-iterative SIAM application was designed to run automatically in near real-time on the web and on mobile devices. Its core is a one-pass prior knowledge-based decision tree for MS reflectance space hyperpolyhedralization into static (non-adaptive-to-data) color names presented in literature in recent years. For the sake of readability this paper is split into two. The present Part 1 – Theory provides the multidisciplinary background of *a priori* color naming in cognitive science, from linguistics to computer vision, and surveys related works on static MS color naming. To cope with dictionaries of MS color names and land cover class names that do not coincide and must be harmonized, an original hybrid (combined deductive and inductive) guideline is proposed to identify a categorical variable-pair relationship. An original quantitative measure of categorical variable-pair association is also proposed. The subsequent Part 2 – Validation presents and discusses Stage 4 *Val* results collected by an original protocol for wall-to-wall thematic map quality assessment without sampling where the test and reference map legends can differ. Conclusions are that the SIAM-WELD maps instantiate a Level 2 SCM product whose legend is the 4-class taxonomy of the Food and Agriculture Organization of the United Nations – Land Cover Classification System (LCCS) at the Dichotomous Phase Level 1 (vegetation/non-vegetation), Level 2 (terrestrial/aquatic) or superior LCCS level.

## 1. Introduction

Proposed by the intergovernmental Group on Earth Observations (GEO), the visionary goal of the Global Earth Observation System of Systems (GEOSS) implementation plan for years 2005-2015 (GEO 2005) is the systematic transformation of multi-source EO "big data" into timely, comprehensive and operational EO value-adding products and services, submitted to the Quality Assurance Framework for Earth Observation (QA4EO) calibration/validation (*Cal/Val*) requirements (GEO-CEOS 2010). To date the GEOSS mission cannot be considered fulfilled by the remote sensing (RS) community. Existing Earth observation (EO) image understanding systems (EO-IUSs) tend to score low in productivity because outpaced by the rate of collection of EO sensory data, whose quality and quantity are ever-increasing. To be considered affected by low levels of productivity, which means to be considered in non-operating mode, an EO-IUS suffices to fall short in one of its outcome and process (OP) quantitative quality indicators ($Q^2$Is), to be community-agreed upon in compliance with the QA4EO guidelines. A proposed minimally dependent and maximally informative (mDMI) set of EO OP-$Q^2$Is includes degree of automation, effectiveness, e.g., thematic mapping accuracy, efficiency in computation time and in memory occupation, robustness (vice versa, sensitivity) to changes in input data, robustness to changes in input parameters to be user-defined, scalability to changes in user requirements and in sensor specifications, timeliness from data acquisition to information product generation, and costs in manpower and computer power (Baraldi and Boschetti 2012a, 2012b; Duke 2016). According to the Pareto formal analysis of multi-objective optimization problems, optimization of an mDMI set of OP-$Q^2$Is is an inherently-ill posed problem in the Hadamard sense (Hadamard 1902), where many Pareto optimal solutions lying on the Pareto efficient frontier can be considered equally good (Boschetti et al. 2004).

The conjecture that existing EO-IUSs are Pareto sub-optimal and tend to score low in operating mode is supported by several facts. First, the percentage of EO data ever downloaded from the European Space Agency (ESA) databases is estimated at about 10% or less (D'Elia 2012). Second, EO-IUSs presented in the RS literature are typically assessed and compared based on the sole mapping accuracy, which means their mDMI set of OP-$Q^2$Is remains largely unknown to date. For example, when a large-scale EO data-derived product was generated by a supervised data learning EO-IUS, the most limiting factors turned out to be the cost, timeliness, quality and availability of adequate supervised (labeled) data samples collected from field sites, existing maps or geospatial data archives in tabular form (Gutman et al. 2004). Third, no ESA EO data-derived Level 2 prototype product has ever been systematically generated at the ground segment (ESA 2015; CNES 2016). An EO Level 2 product is defined by ESA as a multi-spectral (MS) image corrected for geometric, atmospheric, adjacency and topographic effects, equivalent to a multivariate numeric variable of enhanced information quality (related to the concept of quantitative unequivocal *information-as-thing* in the terminology of Capurro and Hjørland [2003]), see Figure 1-1, stacked with its data-derived scene classification map (SCM) (ESA 2015; CNES 2016), equivalent to a categorical variable of semantic quality (related to the concept of qualitative equivocal *information-as-data-interpretation* in the terminology of Capurro and Hjørland [2003]). An EO Level 2 SCM legend is expected to consist of general-purpose, user- and application-independent land cover (LC) classes, in addition to quality layers such cloud and cloud-shadow (ESA 2015; CNES 2016).

A possible example of EO Level 2 SCM legend is the 3-level 8-class Dichotomous Phase (DP) taxonomy of the Food and Agriculture Organization of the United Nations (FAO) - Land Cover Classification System (LCCS) (Di Gregorio and Jansen 2000). The LCCS-DP hierarchy comprises three “nested” dichotomous LC class layers: Level 1 – Vegetation versus non-vegetation, Level 2 – Terrestrial versus aquatic and Level 3 – Managed versus natural or semi-natural. The 3-level 8-class LCCS-DP taxonomy is listed in Figure 1-2. For the sake of generality, a 3-level 8-class LCCS-DP legend added with LC class “other” or “rest of the world”, which includes information layers such as cloud and cloud-shadow, is identified hereafter as “augmented” 9-class LCCS-DP taxonomy. In the complete two-phase LCCS hierarchy a low-level general-purpose LCCS-DP legend is preliminary to a high-level application- and user-specific LCCS Modular Hierarchical Phase (MHP) taxonomy, consisting of a hierarchical battery of one-class LC class-specific classifiers (Di Gregorio and Jansen 2000). In recent years the two-phase LCCS taxonomy has become increasingly popular (Ahlqvist 2008). One reason of its popularity is that the LCCS hierarchy is “fully nested” while alternative LC class hierarchies, such as the CORINE Land Cover (CLC) taxomony (Bossard et al. 2000) and the EO Image Librarian LC taxonomy (Dumitru et al. 2015), start from a Level 1 which is multi-class. In a hierarchical EO-IUS architecture submitted to a garbage in, garbage out information principle, the fully-nested LCCS hierarchy highlights the dependence of OP-$Q^2$Is featured by any high-level LCCS-MHP data processing module on OP-$Q^2$Is featured by low-level LCCS-DP data processing units, starting from the LCCS-DP Level 1 vegetation/non-vegetation information layer whose relevance becomes paramount for all subsequent LCCS layers. This multi-level dependence is neither trivial nor obvious to underline. For example, vegetation/non-vegetation discrimination is acknowledged to be very challenging when pursued in EO image composites at continental or global scale by means of traditional supervised data learning EO-IUSs (Gutman et al. 2004), which are inherently semi-automatic and training data-specific (Liang 2004).

Our work thesis was that a necessary not sufficient pre-condition for a yet-unfulfilled GEOSS development (GEO 2005) is the systematic generation at the ground segment of an ESA EO Level 2 product, whose general-purpose SCM is constrained as follows. First, the SCM legend agrees with the 3-level 9-class “augmented” LCCS-DP taxonomy (see Figure 1-2). Second, to comply with the QA4EO *Cal/Val* requirements the SCM product must be submitted to a GEO Stage 4 *Val*, where an mDMI set of OP-$Q^2$Is is evaluated by independent means (GEO-CEOS 2010). By definition a GEO Stage 3 *Val* requires that “spatial and temporal consistency of the product with similar products are evaluated by independent means over multiple locations and time periods representing global conditions. In Stage 4 *Val*, results for Stage 3 are systematically updated when new product versions are released and as the time-series expands” (GEO-CEOS WGCV 2015).

To contribute toward filling an analytic and pragmatic information gap from multi-source EO big data to the ESA EO Level 2 product, the primary goal of this interdisciplinary study was to undertake an original (to the best of these authors’ knowledge, the first) outcome and process Stage 4 *Val* of an off-the-shelf Satellite Image Automatic Mapper™ (SIAM™) lightweight computer program. Implemented in operating mode in the C/C++ programming language, the SIAM software executable runs: (i) automatically, i.e., it requires no human-machine interaction, (iii) in near real-time, specifically, it is non-iterative (one-pass with a single subsystem that is two-pass, refer to the text below), with a computational complexity increasing linearly with the image size, and (iii) in tile streaming mode, i.e., it requires a fixed runtime memory occupation (Baraldi et al. 2006, 2010a, 2010b, 2010c, 2012a, 2012b, 2013, 2015, 2016; Baraldi and Humber 2015). In addition to running on laptop and desktop computers the SIAM lightweight computer program is eligible for use in a mobile software application. Eventually provided with a mobile user interface, a mobile software application is a lightweight computer program specifically designed to run on web browsers and mobile devices, such as tablet computers and smartphones. The SIAM software pipeline comprises six non-iterative subsystems for MS image analysis (decomposition) and synthesis (reconstruction). Its core is a one-pass prior knowledge-based decision tree for MS reflectance space hyperpolyhedralization into static (non-adaptive-to-data) color names, presented in the RS literature where enough information was provided for the implementation to be reproduced (Baraldi et al. 2006). Sketched in Figure 1-3, the SIAM application workflow is summarized hereafter.

(1) MS data radiometric calibration, in agreement with the QA4EO *Cal* requirements (GEO-CEOS 2010). The SIAM expert system instantiates a physical data model; hence, it requires as input sensory data provided with a physical meaning. Specifically, digital numbers must be radiometrically calibrated into a physical unit of radiometric measure to be community-agreed upon, such as top-of-atmosphere reflectance (TOARF), surface reflectance (SURF) or Kelvin degrees for thermal channels. Relationship TOARF $\supseteq$ SURF holds because SURF is a special case of TOARF in clear sky and flat terrain conditions (Chavez 1988), i.e., TOARF $\approx$ SURF + atmospheric noise + topographic effects + surface adjacency effects. In a spectral decision tree this relationship means that if decision boundaries are able to cope with MS hyperpolyhedra of "noisy" TOARF values then they can also deal with "noiseless" SURF values as special cases of the former, while the vice versa does not necessarily hold, see Figure 1-4.

(2) One-pass prior knowledge-based SIAM decision tree for MS reflectance space hyperpolyhedralization into three static codebooks (dictionaries) of sub-symbolic color names as codewords, see Figure 1-4. Provided with inter-level parent-child relationships, the SIAM's three-level dictionary of static color names features a ColorDictionaryCardinality value which decreases from fine to intermediate to coarse, refer to Table 1-1 and Figure 1-5. MS reflectance space hyperpolyhedra for color naming are difficult to think of and impossible to visualize when the MS data space dimensionality is superior to three. This is not the case of basic color names adopted in human languages (Berlin and Kay 1969), whose mutually exclusive and totally exhaustive perceptual polyhedra, neither necessarily convex nor connected, are intuitive to think of and easy to visualize in a 3D monitor-typical red-green-blue (RGB) cube, see Figure 1-6 (Griffin 2006; Benavente et al. 2008). When each pixel of a MS image is mapped onto a color space partitioned into a set of mutually exclusive and totally exhaustive hyperpolyhedra equivalent to a dictionary of color names, then a 2D multi-level color map is generated automatically (without human-machine interaction) in near real-time (with a computational complexity increasing linearly with the image size), where the number $k$ of 2D map levels (color strata, color names) belongs to range $\{1, \text{ColorDictionaryCardinality}\}$. Popular synonyms of measurement space hyperpolyhedralization (discretization, partition) are vector quantization (VQ) in inductive machine learning-from-data (Cherkassky and Mulier 1998) (Fritzke 1997a, 1997b; Patanè and Russo 2001, 2002; Linde et al. 1980; Lee et al. 1997; Lloyd 1982; Elkan 2003), and deductive fuzzification of a numeric variable into fuzzy sets in fully logic (Zadeh 1965). Typical inductive learning-from-data VQ algorithms aim at minimizing a known VQ error function, e.g., a root mean square error (RMSE), given a number of $k$ discretization levels selected by a user based on *a priori* knowledge and/or heuristic criteria. One of the most widely used VQ heuristics in RS and computer vision (CV) applications is the *k*-means VQ algorithm (Linde et al. 1980; Lee et al. 1997; Lloyd 1982; Elkan 2003), capable of convex Voronoi tessellation of an unlabeled data space (Fritzke 1997a; Cherkassky and Mulier 1998). For example, in a bag-of-words model applied to CV tasks, a numeric color space is typically discretized into a categorical color variable by an inductive VQ algorithm, such as *k*-means; next, the categorical color variable is simplified by a 1st-order histogram representation, which disregards word grammar, semantics and even word-order, but keeps multiplicity; finally, the frequency of each color codeword is used as a feature for training a supervised data learning classifier (Cimpoi et al. 2014). Unlike the *k*-means VQ algorithm where *k* is user-defined and the VQ error is estimated from unlabeled data, a user can fix the target VQ error value, so that it is the free-parameter *k* to be dynamically learned from unlabeled data by an inductive VQ algorithm (Patanè and Russo 2001, 2002), such as ISODATA (Memarsadeghi et al. 2007). It means there is no universal number *k* of static hyperpolyhedra in a vector data space suitable for satisfying any VQ error specification. As a viable strategy to cope with the inherent ill-posedness of VQ problems (Cherkassky and Mulier 1998), the SIAM expert system provides its three pre-defined VQ levels with a per-pixel RMSE estimation required for VQ quality assurance, in compliance with the QA4EO guidelines, refer to point (6) below.

(3) Well-posed (deterministic) two-pass detection of connected-components in the multi-level color map-domain (Dillencourt et al. 1992; Sonka et al. 1994), where the number *k* of map levels is ≤ ColorDictionaryCardinality, see Figure 1-7. These connected-components consist of connected sets of pixels featuring the same color label. They are typically known as superpixels in the CV literature (Achanta et al. 2011), homogeneous segments or image-objects in the object-based image analysis (OBIA) literature (Blaschke et al. 2014; Nagao and Matsuyama 1980; Matsuyama and Hwang 1990; Shackelford and Davis 2003a, 2003b), and texture elements (texels) in human vision (Julesz 1986; Julesz et al. 1973). Whereas the physical model-based SIAM expert system requires no human-machine interaction to detect top-down superpixels whose shape and size can be any, superpixels detected bottom-up in CV applications typically require a pair of statistical model's free-parameters to be user-defined based on heuristics. This pair of user-defined parameters typically thresholds the superpixel maximum area and forces a superpixel to stay compact in shape (Achanta et al. 2011). In a multi-level image domain where *k* is the number of levels (image-wide strata), superpixels, pixels and strata co-exist as labeled spatial units provided with a parent-child relationship, where each superpixel is a 2-tuple [superpixel ID, level 1-of-*k*] and each pixel is a 2-tuple [raw-column coordinate pair, superpixel ID].

(4) Well-posed 4- or 8-adjacency cross-aura representation in linear time of superpixel-contours, see Figure 1-8. These cross-aura contour values allow estimation of a scale-invariant planar shape index of compactness (Soares et al. 2014), eligible for use by a high-level OBIA approach (Blaschke et al. 2014), see Figure 1-3.

(5) Superpixel description table allocation and initialization, to describe superpixels in a 1D tabular form (list) in combination with their 2D raster representation, to take advantage of each data structure and overcome their shortcomings (Nagao and Matsuyama 1980; Matsuyama and Hwang 1990; Marr 1980). Typically local spatial searches are computational more efficient in the raster domain than in the list representation (Marr 1980).

(6) Superpixelwise-constant input image approximation (reconstruction), also known as "image-object mean view" in OBIA applications (Trimble 2015), followed by a per-pixel RMSE estimation between the original MS image and the reconstructed piecewise-constant MS image. This VQ error estimation strategy enforces a product quality assurance policy considered mandatory by the QA4EO guidelines. For example, VQ quality assurance supported by SIAM allows a user to adopt quantitative (objective) criteria in the selection of pre-defined VQ levels to fit user- and application-specific VQ error requirement specifications.

An example of the SIAM output products automatically generated in linear time from a radiometrically calibrated 13-band 10 m-resolution Sentinel-2A image is shown in Figure 1-9.

The potential impact on the RS community of a Stage 4 *Val* of an off-the-shelf SIAM lightweight computer program for prior knowledge-based MS reflectance space hyperpolyhedralization, superpixel detection and per-pixel VQ quality assessment is expected to be relevant, with special emphasis on existing or future *hybrid* (combined deductive and inductive) EO-IUSs. In the RS discipline there is a long history of prior knowledge-based MS reflectance space partitioners for static color naming, alternative to SIAM's, developed but never validated by space agencies, public organizations and private companies for use in hybrid EO-IUSs in operating mode, see Figure 1-10. Examples of hybrid EO data pre-processing applications (*information-as-thing*) conditioned by static color naming are large-scale MS image compositing (Ackerman et al. 1998; Luo et al. 2008; Lück and van Niekerk 2016), MS image atmospheric correction (Richter and D. Schläpfer 2012a; Richter and D. Schläpfer 2012b; Baraldi, Humber and Boschetti 2013; Baraldi and Humber 2015; Dorigo et al. 2009; Vermote and Saleous 2007; DLR and VEGA 2011; Lück and van Niekerk 2016), MS image topographic correction (Richter and D. Schläpfer 2012a; Richter and D. Schläpfer 2012b; Baraldi, Humber and Boschetti 2013; Baraldi and Humber 2015; Dorigo et al. 2009; Baraldi et al. 2010c; DLR and VEGA 2011; Lück and van Niekerk 2016), see Figure 1-11, MS image adjacency effect correction (DLR and VEGA 2011) and radiometric quality assurance of pan-sharpened MS imagery (Despini et al. 2014). Examples of hybrid EO image classification applications (*information-as-data-interpretation*) conditioned by static color naming are cloud and cloud-shadow quality layer detection (Baraldi et al. 2015; DLR and VEGA 2011; Lück and van Niekerk 2016), single-date LC classification (Muirhead and Malkawi 1989; Simonetti et al. 2015a; GeoTerraImage 2015; DLR and VEGA 2011; Lück and van Niekerk 2016), multi-temporal post-classification LC change (LCC)/no-change detection (Baraldi et al. 2016; Tiede et al. 2016; Simonetti et al. 2015a), multi-temporal vegetation gradient detection and quantization in fuzzy sets (Arvor et al. 2016), multi-temporal burned area detection (Boschetti et al. 2015), and prior knowledge-based LC mask refinement (cleaning) of supervised data samples employed as input to supervised data learning EO-IUSs (Baraldi et al. 2010a, 2010b). Due to their large application domain ranging from low- (pre-attentional) to high-level (attentional) vision tasks, existing hybrid EO-IUSs in operating mode conditioned by static color naming are natural candidates for the systematic transformation of multi-source single-date MS imagery into EO Level 2 product at the ground segment.

The terminology adopted in the rest of this paper is mainly driven from the multidisciplinary domain of cognitive science, see Figure 1-12. Popular synonyms of deductive inference are top-down, prior knowledge-based, learning-from-rule and physical model-based inference. Synonyms of inductive inference are bottom-up, learning-from-data, learning-from-examples and statistical model-based inference (Baraldi and Boschetti 2012a, 2012b; Liang 2004). Hybrid inference systems combine statistical and physical models to take advantage of the unique features of each and overcome their shortcomings (Baraldi and Boschetti 2012a, 2012b; Cherkassky and Mulier 1998; Liang 2004). For example, in biological cognitive systems "there is never an absolute beginning" (Piaget 1970), where an *a priori* genotype provides initial conditions to an inductive learning-from-examples phenotype (Parisi 1991). Biological cognitive systems are hybrid inference systems where inductive/phenotypic learning-from-examples mechanisms explore the neighbourhood of deductive/genotypic initial conditions in a solution space (Parisi 1991). In line with biological cognitive systems an artificial hybrid inference system can alternate deductive and inductive inference units, starting from a deductive first stage for initialization purposes, see Figure 1-10. It means that no deductive inference subsystem, such as SIAM, should be considered stand-alone, but eligible for use in a hybrid inference system architecture to initialize (pre-condition, stratify) inductive learning-from-data algorithms, which are inherently ill-posed (difficult-to-solve) and require *a priori* knowledge in addition to data to become better posed for numerical solution (Cherkassky and Mulier 1998).

To comply with the GEO Stage 4 *Cal*/*Val* requirements, the selected ready-for-use SIAM application had to be validated by independent means on a radiometrically calibrated EO image time-series at large spatial extent. This input data set was identified in the open-access U.S. Geological Survey (USGS) 30 m resolution Web Enabled Landsat Data (WELD) annual composites of the conterminous U.S. (CONUS) for the years 2006 to 2009, radiometrically *Cal* into TOARF values (Roy et al. 2010; Homer et al. 2004; WELD 2015). The 30 m resolution 16-class U.S. National Land Cover Data (NLCD) 2006 map, delivered in 2011 by the U.S. Geological Survey (USGS) Earth Resources Observation Systems (EROS) Data Center (EDC) (Vogelmann et al. 1998, 2001; Wickham et al. 2010; Wickham et al. 2013; Xian and Homer 2010; EPA 2007), was selected as the reference thematic map at continental spatial extent. The 16-class NLCD map legend is summarized in Table 1-2. To account for typical non-stationary geospatial statistics, the NLCD 2006 thematic map was partitioned into 86 Level III ecoregions of North America collected from the Environmental Protection Agency (EPA) (EPA 2013; Griffith and Omernik 2009).

In this experimental framework the test SIAM-WELD annual color map time-series and the reference NLCD 2006 map share the same spatial extent and spatial resolution, but their map legends are not the same. These working hypotheses are neither trivial nor conventional in the RS literature where thematic map quality assessment strategies typically adopt an either random or non-random sampling strategy and assume that the test and reference thematic map dictionaries coincide (Stehman and Czaplewski 1998). Starting from a stratified random sampling protocol presented in (Baraldi et al. 2014), the secondary contribution of the present study was to develop a novel protocol for wall-to-wall comparison without sampling of two thematic maps featuring the same spatial extent and spatial resolution, but whose legends can differ.

For the sake of readability this paper is split into two, the present Part 1 – Theory and the subsequent Part 2 – Validation. An expert reader familiar with static color naming in cognitive science, spanning from linguistics to human vision and computer vision, can skip the present Part 1. To make this paper self-contained and provided with a relevant survey value, the Part 1 is organized as follows. The multidisciplinary background of color naming is discussed in Section 2. Section 3 reviews prior knowledge-based decision trees for MS color naming presented in the RS literature. To cope with thematic map legends that do not coincide and must be harmonized (reconciled, associated, translated) (Ahlqvist 2005), such as dictionaries of MS color names and LC class names, Section 3 proposes an original hybrid guideline to identify a categorical variable-pair relationship, where prior beliefs are combined with additional evidence inferred from new data. An original measure of categorical variable-pair association is proposed in Section 4. In the subsequent Part 2 Stage 4 *Val* results are collected by an original protocol for wall-to-wall thematic map quality assessment without sampling where the test SIAM-WELD map legend and the reference NLCD 2006 map legend are harmonized. Conclusions are that the SIAM-WELD maps instantiate an EO Level 2 SCM product whose legend is the FAO LCCS taxonomy at the DP Level 1 (vegetation/non-vegetation), Level 2 (terrestrial/aquatic) or superior LCCS level.

## 2. Color naming problem background in cognitive science

Vision is an inherently ill-posed cognitive problem where scene-from-image representation is affected, first, by data dimensionality reduction from the 4D spatiotemporal scene-domain to the (2D) image-domain and, second, by a semantic information gap from ever-varying sub-symbolic numeric sensations to stable categorical and semantic (symbolic) percepts (Matsuyama and Hwang 1990). In vision, spatial topological and spatial non-topological information typically dominate color information. This thesis is proved by the undisputable fact that achromatic (panchromatic) human vision, familiar to everybody when wearing sunglasses, is nearly as effective as chromatic vision in scene-from-image representation. It means that a necessary not sufficient condition for a CV system to fully exploit spatial topological and non-topological information in addition to color is to perform nearly the same when input with panchromatic or color imagery.

Deeply investigated in CV (Sonka et al. 1994; Frintrop 2011), content-based image retrieval (Smeulders et al. 2000) and RS applications (Baraldi and Boschetti 2012a, 2012b; Nagao and Matsuyama 1980; Matsuyama and Hwang 1990; Shackelford and Davis 2003a, 2003b), popular visual features are: (i) color (Griffin 2006; Gevers et al. 2012; ii) local shape (Wenwen Li et al. 2013; iii) texture, defined as the perceptual spatial grouping of texture elements known as texels (Julesz 1986; Julesz et al. 1973) or tokens (Marr 1982; iv) inter-object spatial topological relationships, e.g., adjacency, inclusion, etc., and (v) inter-object spatial non-topological relationships, e.g., spatial distance, angle measure, etc. Color is the sole visual property available at the imaging sensor resolution. In other words, pixel-based information is spatial context-independent and purely color-specific. Among the aforementioned visual variables, per-pixel color values are the sole non-spatial numeric variable.

Neglecting the fact that spatial topological and non-topological information typically dominate color information in both the (2D) image-domain and the 4D spatiotemporal scene-domain involved with vision (Matsuyama and Hwang 1990), traditional EO-IUSs adopt a 1D image analysis approach, see Figure 1-13. In 1D image analysis, a 1D streamline of vector data, either spatial context-sensitive (e.g., window-based or image object-based) or context-insensitive (pixel-based), is processed irrespective of the order of presentation of the input sequence. In practice 1D image analysis is invariant to permutations, such as in orderless encoders (Cimpoi et al. 2014). When vector data are spatial context-sensitive then 1D image analysis ignores spatial topological information. When vector data are pixel-based then 1D image analysis ignores both spatial topological and non-topological information. Prior knowledge-based color naming of a spatial unit x in the image-domain, where x is either (0D) point, (1D) line or (2D) polygon defined according to the Open Geospatial Consortium (OGC) nomenclature (OGC 2015), is a special case of 1D image analysis, either pixel-based or image object-based, where spatial topological and/or non-topological information are ignored.

Alternative to 1D image analysis, 2D image analysis relies on a sparse (distributed) 2D array (2D regular grid) of local spatial filters (Tsotsos 1990), suitable for topology-preserving feature mapping (Martinetz et al. 1994; Fritzke 1997a), see Figure 1-14. The human brain's organizing principle is topology-preserving feature mapping (Feldman 2016). In the biological visual system, topology-preserving feature maps are primarily spatial, where activation domains of physically adjacent processing units in the 2D array of convolutional filters are spatially adjacent regions in the 2D visual field. Provided with a superior degree of biological plausibility in modelling 2D spatial topological and non-topological information, distributed processing systems capable of 2D image analysis, such as deep convolutional neural networks (DCNNs), outperform 1D image analysis approaches (Cimpoi et al. 2014). This apparently trivial consideration is at odd with a relevant portion of the RS literature, where pixel-based 1D image analysis is mainstream followed by context-sensitive 1D image analysis implemented within the OBIA paradigm (Blaschke et al. 2014).

Since traditional EO-IUSs adopt a 1D image analysis approach where dominant spatial information is neglected in favour of secondary color information, it is useful to turn attention to the multidisciplinary framework of cognitive science to shed light on how humans deal with color information. According to cognitive science, which includes linguistics, the study of languages, humans discretize (fuzzify) ever-varying quantitative (numeric) photometric and spatiotemporal sensations into stable qualitative (categorical, nominal) percepts, eligible for use in symbolic human reasoning based on a convergence-of-evidence approach (Matsuyama and Hwang 1990). In their seminal work, Berlin and Kay proved that 20 human languages, spoken across space and time in the real-world, partition quantitative color sensations collected in the visible portion of the electromagnetic spectrum (see Figure 1-1) onto the same "universal" dictionary of eleven basic color (BC) names (Berlin and Kay 1969): black, white, gray, red, orange, yellow, green, blue, purple, pink and brown. In a 3D monitor-typical red-green-blue (RGB) cube, BC names are intuitive to think of and easy to visualize. They provide a mutually exclusive and totally exhaustive partition of the RGB cube into RGB polyhedra neither necessarily connected nor convex, see Figure 1-6 (Griffin 2006; Benavente et al. 2008). Since they are community-agreed upon to be used by members of the community, RGB BC polyhedra are prior knowledge-based, i.e., stereotyped, non-adaptive-to-data (static), general-purpose, application- and data-independent. Multivariate measurement space hyperpolyhedralization is the transformation of a numeric variable into a categorical variable. This is a typical problem in many scientific disciplines, such as inductive VQ in machine learning-from-data (Cherkassky and Mulier 1998) and deductive numeric variable fuzzification into discrete fuzzy sets in fuzzy logic (Zadeh 1965), refer to Section 1.

In an analytic model of vision based on a convergence-of-evidence approach, the first original contribution of the present Part 1 is to intuitively show how prior knowledge-based color value discretization into a static dictionary of color names affects image classification. If individual sources of visual evidence, such as color, local shape, texture and inter-object spatial relationships, are considered statistically independent, when target classes of observed objects in the real-world scene are c = 1, …, ObjectClassLegendCardinality, then for a given discrete spatial unit x in the image-domain, either point, line or polygon (OGC 2015), it is possible to write, based on the Bayesian law

p(c| ColorValue(x), ShapeValue(x), TextureValue(x), SpatialRelationships(x, Neigh(x))) =
p(c, ColorValue(x), ShapeValue(x), TextureValue(x), SpatialRelationships(x, Neigh(x))) /
p(ColorValue(x), ShapeValue(x), TextureValue(x), SpatialRelationships(x, Neigh(x))) =
p(c| ColorValue(x)) • p(c| ShapeValue(x)) • p(c| TextureValue(x)) • p(c| SpatialRelationships(x, Neigh(x))) =
[p(ColorValue(x)| c) • p(c) / p(ColorValue(x))] • [p(ShapeValue(x)| c) • p(c) / p(ShapeValue(x))] • [p(TextureValue(x)| c) • p(c) / p(TextureValue(x))] •
[p(SpatialRelationships(x, Neigh(x))| c) • p(c) / p(SpatialRelationships(x, Neigh(x)))] ≤
min{p(c| ColorValue(x)), p(c| ShapeValue(x)), p(c| TextureValue(x)), p(c| SpatialRelationships(x, Neigh(x)))}, c = 1, …, ObjectClassLegendCardinality, (1-1)

where ColorValue(x) belongs to a MS measurement space $\Re^{MS}$ and Neigh(x) is a generic 2D spatial neighborhood of spatial unit x in the image-domain. Equation (1-1) shows that any convergence-of-evidence approach is more selective than each individual source of evidence, in line with a focus-of-visual attention mechanism (Frintrop 2011). For the sake of simplicity, if priors are ignored because considered equiprobable in a maximum class-conditional likelihood inference approach alternative to a maximum *a posteriori* optimization criterion, then Equation (1-1) becomes

p(c| ColorValue(x), ShapeValue(x), TextureValue(x), SpatialRelationships(x, Neigh(x))) ∝
[p(ColorValue(x)| c) / p(ColorValue(x))] • [p(ShapeValue(x)| c) / p(ShapeValue(x))] •
[p(TextureValue(x)| c) / p(TextureValue(x))] • [p(SpatialRelationships(x, Neigh(x))| c) / p(SpatialRelationships(x, Neigh(x)))] =
[$\sum_{\text{ColorName}=1}^{\text{ColorDictionaryCardinality}} p(\text{ColorValue}(x)|\text{ColorName})p(\text{ColorName}|c)$ / p(ColorValue(x))] • [p(ShapeValue(x)| c) / p(ShapeValue(x))] • [p(TextureValue(x)| c) / p(TextureValue(x))] • [p(SpatialRelationships(x, Neigh(x))| c) / p(SpatialRelationships(x, Neigh(x)))], c = 1, …, ObjectClassLegendCardinality, (1-2)

where color space $\Re^{MS}$ is partitioned into hyperpolyhedra, equivalent to a discrete and finite dictionary of static color names, with ColorName = 1, .., ColorDictionaryCardinality. To further simplify Equation (1-2), its canonical interpretation based on frequentist statistics can be relaxed by fuzzy logic (Zadeh 1965), so that the logical-AND operator is replaced by a fuzzy-AND (min) operator, inductive class-conditional probability p(x| c) ∈ [0, 1], where $\sum_{c=1}^{\text{ObjectClassLegendCardinality}} p(x|c) \geq 0$, is replaced by a deductive membership (compatibility) function m(x| c) ∈ [0, 1], where $\sum_{c=1}^{\text{ObjectClassLegendCardinality}} m(x|c) \geq 0$, and color space hyperpolyhedra are considered mutually exclusive and totally exhaustive. If these simplifications are adopted, then Equation (1-2) becomes

m(c| ColorValue(x), ShapeValue(x), TextureValue(x), SpatialRelationships(x, Neigh(x))) ∝
min{ $\sum_{\text{ColorName}=1}^{\text{ColorDictionaryCardinality}} m(\text{ColorValue}(x)|\text{ColorName})m(\text{ColorName}|c)$, m(ShapeValue(x)| c), m(TextureValue(x)| c), m(SpatialRelationships(x, Neigh(x))| c)} =
min{m(ColorName*| c), m(ShapeValue(x)| c), m(TextureValue(x)| c), m(SpatialRelationships(x, Neigh(x))| c)}, c = 1, …, ObjectClassLegendCardinality, where

ColorName* ∈ {1, ColorDictionaryCardinality}, such that m(ColorValue(x)| ColorName*) = 1 and m(ColorName*| c) ∈ {0, 1}. (1-3)

In Equation (1-3), the following considerations hold.

- Each numeric ColorValue(x) in color space $\Re^{MS}$ belongs to a single color name (hyperpolyhedron) ColorName* in the static color dictionary, i.e., ∀ ColorValue(x) ∈ $\Re^{MS}$, $\sum_{\mathrm{ColorName}=1}^{\mathrm{ColorDictionaryCardinality}} \mathrm{m(ColorValue(x)|ColorName)} = \mathrm{m(ColorValue(x)|ColorName^{*})} = 1$ holds, where m(ColorValue(x)| ColorName) ∈ {0, 1}, ColorName = 1, …, ColorDictionaryCardinality.
- The set A = DictionaryOfColorNames, with cardinality |A| = a = ColorDictionaryCardinality, and the set B = LegendOfObjectClassNames, with cardinality |B| = b = ObjectClassLegendCardinality, can be considered a bivariate categorical random variable where two univariate categorical variables A and B are generated from a single population. A binary relationship (product set) from set A to set B, R: A ⇒ B, is a subset of the 2-fold Cartesian product A × B, whose size is rows × columns = a × b. The Cartesian product of two sets A × B is a set whose elements are ordered pairs. Hence, the Cartesian product is non-commutative, A × B ≠ B × A. In agreement with common sense, see Table 1-3, R: DictionaryOfColorNames ⇒ LegendOfObjectClassNames is a set of ordered pairs where each ColorName can be assigned to none, one or several classes c = 1, …, ObjectClassLegendCardinality of observed scene-objects, whereas each class of observed objects can be assigned with none, one or several color names as the class-specific color attribute. Binary membership values m(ColorName| c) ∈ {0, 1} and m(c| ColorName) ∈ {0, 1}, with c = 1, …, ObjectClassLegendCardinality and ColorName = 1, …, ColorDictionaryCardinality, can be community-agreed upon based on various kinds of evidence, whether viewed all at once or over time, such as a combination of prior beliefs with additional evidence inferred from new data in agreement with a Bayesian updating rule (Bayesian inference), largely applied in artificial intelligence and expert systems. A binary relationship R: A ⇒ B ⊆ A × B where sets A and B are categorical variables generated from a single population guides the interpretation process of a two-way *contingency table* (also known as association matrix, cross tabulation, bivariate table or frequency table) (Kuzera and Pontius 2008; Pontius and Connors 2006), BIVRTAB = FrequencyCount(A × B). In the conventional domain of frequentist inference with no reference to prior beliefs, a BIVRTAB is the 2-fold Cartesian product A × B instantiated by the bivariate frequency counts of the two univariate categorical variables A and B generated from a single population. For any BIVRTAB instance, either square or non-square, there is a binary relationship R: A ⇒ B ⊆ A × B that guides the interpretation process, where "correct" entry-pair cells of the 2-fold Cartesian product A × B can be either off-diagonal (scattered) or on-diagonal, if a main diagonal exists. When a BIVRTAB is estimated from a geospatial population, it is called *overlapping area matrix* (OAMTRX) (Baraldi et al. 2014; Baraldi et al. 2006; Lunetta and Elvidge 1999; Beauchemin and Thomson 1997; Ortiz and Oliver 2006; Baraldi et al. 2005; Pontius and Connors 2006). When the binary relationship R: A ⇒ B is a bijective function (both 1-1 and onto), i.e., when the two categorical variables A and B estimated from a single population coincide, then the BIVRTAB is square and sorted and typically called confusion matrix (CMTRX) or error matrix

(Stehman and Czaplewski 1998; Congalton and Green 1999; Lunetta and Elvidge 1999). In a CMTRX the main diagonal guides the interpretation process. For example, a square OAMTRX = FrequencyCount(A × B), where A = test thematic map legend, B = reference thematic map legend and cardinality a = b, is a CMTRX if and only if A = B, i.e., if the test and reference codebooks are the same sorted set of semantic concepts. In general the class of (square and sorted) CMTRX instances is a special case of the class of OAMTRX instances, either square or non-square, i.e., OAMTRX ⊃ CMTRX. A similar consideration holds about summary $Q^2$Is generated from an OAMTRX or a CMTRX, i.e., $Q^2$I(OAMTRX) ⊃ $Q^2$I(CMTRX).

Equation (1-3) shows that for any spatial unit x in the image-domain, when a hierarchical CV classification approach estimates posterior m(c| ColorValue(x), ShapeValue(x), TextureValue(x), SpatialRelationships(x, Neigh(x))) starting from a near real-time context-insensitive color naming first stage where condition m(ColorValue(x)| ColorName*) = 1 holds, if condition m(ColorName*| c) = 0 is true according to a static community-agreed binary relationship R: DictionaryOfColorNames ⇒ LegendOfObjectClassNames (and vice versa) known *a priori*, see Table 1-3, then m(c| ColorValue(x), ShapeValue(x), TextureValue(x), SpatialRelationships(x, Neigh(x))) = 0 irrespective of any second-stage assessment of spatial terms ShapeValue(x), TextureValue(x) and SpatialRelationships(x, Neigh(x)), whose computational model is typically difficult to find and computationally expensive. Intuitively Equation (1-3) shows that static color naming allows the stratification of unconditional multivariate spatial variables into color class-conditional data distributions, in agreement with the statistic stratification principle (Hunt and Tyrrell 2012) and the divide-and-conquer problem solving approach (Bishop 1995; Cherkassky and Mulier 1998). Well known in statistics, the principle of statistic stratification guarantees that "stratification will always achieve greater precision provided that the strata have been chosen so that members of the same stratum are as similar as possible in respect of the characteristic of interest" (Hunt and Tyrrell 2012).

Whereas color polyhedra are easy to visualize and intuitive to think of in a true- or false-color RGB cube, see Figure 1-6, hyperpolyhedra are difficult to think of and impossible to visualize in a MS reflectance space whose dimensionality is superior to three, with spectral channels ranging from visible to thermal portions of the electromagnetic spectrum, see Figure 1-1. Since it is non-adaptive-to-data, any static hyperpolyhedralization of a MS measurement space must be based on *a priori* physical knowledge available in addition to sensory data. Since it relies on a physical data model, static hyperpolyhedralization of a MS data space requires all spectral channels to be provided with a physical unit of radiometric measure, i.e., MS data must be radiometrically calibrated in compliance with the QA4EO *Cal* requirements, refer to Section 1.

Noteworthy, sensory data provided with a physical unit of measure can be input to statistical and physical models, including hybrid inference systems, refer to Section 1. On the contrary, uncalibrated dimensionless sensory data can be input to statistical data models exclusively.

Although considered mandatory by the QA4EO guidelines and regarded as a well-known "prerequisite for physical model-based analysis of airborne and satellite sensor measurements in the optical domain" (Schaepman-Strub et al. 2006, 29), EO data *Cal* is ignored by a relevant portion of the existing RS literature (Baraldi 2009). One consequence is that, to date, statistical model-based EO-IUSs dominate the RS literature and commercial EO image processing software toolboxes, which typically consist of overly complicated collections of inherently ill-posed inductive machine learning-from-data algorithms (Cherkassky and Mulier 1994) to choose from based on heuristics (Baraldi and Boschetti 2012a, 2012b).

## 3. Related works in static MS reflectance space hyperpolyhedralization

In the RS discipline there is a long history of hybrid EO-IUSs in operating mode, suitable for either low-level EO image enhancement or high-level EO image classification, where an *a priori* knowledge-based decision tree for static MS reflectance space hyperpolyhedralization is plugged-in without validation, refer to Section 1. For example, the SIAM stratification of single-date MS imagery contributed to make MS image topographic correction, which is a traditional chicken-and-egg dilemma, better conditioned for automated solution (Baraldi et al. 2010c), in compliance with the EO Level 2 product requirements (ESA 2015; CNES 2016), see Figure 1-11. In the Atmospheric/Topographic Correction for Satellite Imagery (ATCOR) commercial software product, several deductive decision trees are implemented for use in different stages of an EO data enhancement pipeline (Richter and D. Schläpfer 2012a; Richter and D. Schläpfer 2012b; Baraldi, Humber and Boschetti 2013; Baraldi and Humber 2015; Dorigo et al. 2009; Richter et al. 2009), see Figure 1-10. One of the ATCOR's prior knowledge-based per-pixel decision trees delivers as output a haze/ cloud/ water (and snow) classification mask file ("image_out_hcw.bsq"). In addition ATCOR includes a so-called prior knowledge-based Spectral Classification of surface reflectance signatures (SPECL) (Baraldi, Humber and Boschetti 2013; Baraldi and Humber 2015; Dorigo et al. 2009), see Table 1-4. Unfortunately, SPECL has never been tested by its authors in the RS literature, although it has been validated by independent means (Baraldi, Humber and Boschetti 2013; Baraldi and Humber 2015). Supported by the National Aeronautics and Space Administration (NASA), atmospheric effect removal by the Landsat Ecosystem Disturbance Adaptive Processing System (LEDAPS) project relies on exclusion masks for water, cloud, shadow and snow surface types detected by a simple set of prior knowledge-based spectral decision rules. Unfortunately, quantitative analyses of LEDAPS products revealed that these exclusion masks are prone to errors, to be corrected in future LEDAPS releases (Vermote and Saleous 2007). In the 1980s, to provide an automatic alternative to a visual and subjective assessment of the cloud cover on Advanced Very High Resolution Radiometer (AVHRR) quicklook images in the European Space Agency (ESA) Earthnet archive, Muirhead and Malkawi developed a simple algorithm to classify daylight AVHRR images on a pixel-by-pixel basis into land, cloud, sea, snow or ice and sunglint, such that the classified quicklook image was presented in appropriate pseudo-colors, e.g., green: land, blue: sea, white: cloud, etc. (Muirhead and Malkawi 1989). Developed independently by NASA (Ackerman et al. 1998) and the Canadian Center for Remote Sensing (Luo et al. 2008), pixel-based static decision trees contribute, to date, to the generation of clear-sky Moderate Resolution Imaging Spectroradiometer (MODIS) composites, see Figure 1-15. To pursue high-level LC/LCC detection through time, extensions to the time domain of a single-date *a priori* spectral rule base for MS reflectance space quantization have become available to the general public in 2015 through the Google Earth Engine (GEE) platform (Simonetti et al. 2015b) or in the form of a commercial LC/LCC map product at national scale (GeoTerraImage 2015). These two multi-temporal spectral decision trees for LC/LCC detection share the same operational limitations, specifically, they are Landsat sensor series-specific and pixel-based. They were both inspired by a year 2006 SIAM instantiation of a static decision tree for Landsat reflectance space hyperpolyhedralization, presented in pseudo-code in the RS literature (Baraldi et al. 2006) and further developed into the SIAM application software available to date (Baraldi, Humber and Boschetti 2013; Baraldi and Humber 2015; Baraldi and Boschetti 2012a, 2012b; Baraldi et al. 2014; Baraldi et al. 2010a, 2010b; Baraldi 2011; Baraldi et al. 2010d). In Boschetti et al. (2015), a year 2013 SIAM application was successfully employed to accomplish burned area detection in MS image time-series. Among the aforementioned static decision trees for MS color naming, only SIAM claims scalability to several families of EO imaging sensors featuring different spectral resolutions, see Table 1-1.

It is obvious but not trivial to emphasize to the RS community that, in human vision and CV, an *a priori* dictionary of general-purpose data- and application-independent color names is equivalent to a static sub-symbolic categorical variable non-coincident with a symbolic categorical variable whose levels are user- and application-specific classes of objects observed in the 4D spatiotemporal scene-domain, refer to Table 1-3 and Equation (1-3). The very same consideration holds for any discrete and finite set of spectral endmembers in mixed pixel analysis, which "cannot always be inverted to unique LC class names" (Adams et al. 1995, 147).

Quite surprisingly, the non-coincident assumption between an *a priori* dictionary of sub-symbolic color names and a scene- and application-specific legend of symbolic classes of real-world objects appears somehow difficult to acknowledge by relevant portions of the RS community. For example, in the DigitalGlobe Geospatial Big Data platform (GBDX), a patented prior knowledge-based decision tree for pixel-based very high resolution WorldView-2 and WorldView-3 image mapping onto static MS reflectance space hyperpolyhedra (GBDX Registered Name: protogenV2LULC, Provider: GBDX) was proposed to RS end-users as an "Automated Land Cover Classification" (DigitalGlobe 2016). This program name can be considered somehow misleading because assigned to a static sub-symbolic color space partitioner, see Equation (1-3). In fact, so-called "known issues" linked to the "Automated Land Cover Classification" computer program included: "Vegetation: Thin cloud (cloud edges) might be misinterpreted as vegetation; Water: False positives maybe present due to certain types of concrete roofs or shadows; Soils: Ceramic roofing material and some types of asphalt may be misinterpreted as soil," etc. (DigitalGlobe 2016). In Salmon et al. (2013), a year 2006 SIAM's *a priori* dictionary of static sub-symbolic MS color names was downscaled in cardinality and sorted in the order of presentation to form a bijective function with a legend of symbolic classes of target objects in the scene-domain. In practice these authors forced a non-square BIVRTAB to become a CMTRX, where the main diagonal guides the interpretation process (Congalton and Green 1999), to make it more intuitive and familiar to RS practitioners. In general, no binary relationship R: A $\Rightarrow$ B between an *a priori* dictionary A of static sub-symbolic color names and a user- and application-dependent dictionary B of symbolic classes of observed objects in the scene-domain is a bijective function, refer to Table 1-3 and Equation (1-3). As a consequence of its unrealistic hypothesis in color information/knowledge representation, the 1D image classification approach proposed in (Salmon et al. 2013) scored low in accuracy. Unfortunately, to explain their poor MS image classification outcome these authors concluded that, in their experiments, a year 2006 SIAM's static dictionary of color names was useless to identify target LC classes. The lesson to be gained by these authors' experience is that well-established RS practices, such as 1D image analysis based on supervised data learning algorithms and thematic map quality assessment by means of a CMTRX where test and reference thematic legends are the same, can become malpractices when an *a priori* dictionary of static color names is employed for MS image classification purposes in agreement with Equation (1-3) and common sense, see Table 1-3. This lesson learned is supported by the fact that one of the same co-authors of paper (Salmon et al. 2013) reached opposite conclusions when a year 2013 SIAM application software, the same investigated by the present paper, was employed successfully in detecting burned areas from MS image time-series according to a convergence of color names with spatiotemporal visual properties in agreement with Equation (1-3) (Boschetti et al. 2015).

## 4. Original hybrid eight-step guideline for identification of a categorical variable-pair relationship

Our experimental project required to compare an annual time-series of test SIAM-WELD maps of sub-symbolic color names, see Figure 1-5, with a reference NLCD 2006 map whose legend of symbolic LC classes is summarized in Table 1-2. Since these test and reference map legends do not coincide, they must be reconciled through a binary relationship R: DictionaryOfColorNames $\Rightarrow$ LegendOfObjectClassNames (and vice versa), refer to Equation (1-3).

The harmonization of ontologies and the comparison of thematic maps with different legends are the subject of research of a minor body of literature, e.g., refer to works in ontology-driven geographic information systems (ODGIS) (Fonseca et al. 2002; Guarino 1995; Sowa 2000). Ahlqvist writes that “to negotiate and compare information stemming from different classification systems (Bishr 1998; Mizen et al. 2005)... a translation can be achieved by *matching the concepts in one system with concepts in another*, either directly or through an intermediate classification (Feng and Flewelling 2004; Kavouras and Kokla 2002)” (Ahlqvist 2005). Stehman describes four common types of map-pair comparisons (Stehman 1999). In the first type, different thematic maps, either crisp of fuzzy, of the same region of interest and employing the same sorted set (legend) of LC classes are compared (Kuzera and Pontius 2008). In the second type, which includes the first type as a special case, thematic maps, either crisp of fuzzy, of the same region of interest, but featuring map legends that differ in their basic terms with regard to semantics and/or cardinality and/or order of presentation are compared. The third and fourth types of thematic maps comparison regard maps of different surface areas featuring, respectively, the same dictionary or different dictionaries of basic terms. Whereas a large portion of the RS community appears concerned with the aforementioned first type of map comparisons exclusively, the protocol proposed in (Baraldi et al. 2014) is focused on the second type, which includes the first type as a special case. In (Couclelis 2010) Couclelis observed that inter-dictionary concept matching (“conceptual matching” (Ahlqvist 2005)) is an inherently equivocal *information-as-data-interpretation* process (Capurro and Hjørland 2003), see Table 1-3. In common practice two independent human domain-experts (cognitive agents, knowledge engineers [Laurini and Thompson 1992]) are likely to identify different binary associations between two codebooks of codewords. The conclusion is that no "universal best match" of two different codebooks can exist, but identification of the most appropriate binary relationship between two different nomenclatures becomes a subjective matter of negotiation to become community-agreed upon (Couclelis 2010; Capurro and Hjørland 2003).

To streamline the inherently subjective selection of "correct" entry-pairs in a binary relationship R: A $\Rightarrow$ B $\subseteq$ A $\times$ B between two univariate categorical variables A and B of a single population, an original hybrid eight-step guideline was designed for best practice, where deductive/top-down prior beliefs and inductive/bottom-up learning-from-data inference are combined. This hybrid protocol is sketched hereafter as the second original and pragmatic contribution of the present Part 1 to fill the gap from EO sensory data to EO Level 2 product. As an example, let us consider a binary relationship R: A ⇒ B = DictionaryOfColorNames ⇒ LegendOfObjectClassNames ⊆ A × B where rows are a test set of three semi-symbolic color names, say, A = {MS green-as-"*Vegetation*", MS white-as-"*Cloud*", "*Unknowns*"}, where |A| = a = ColorDictionaryCardinality = *TC* = 3 is the row (test) cardinality, and where columns are a reference set of three symbolic LC classes, say, B = {"*Evergreen Forest*", "*Deciduous Forest*", "*Others*"}, where |B| = b = ObjectClassLegendCardinality = *RC* = 3 is the column (reference) cardinality.

1. Display multivariate frequency distributions of the two univariate categorical variables estimated from a single population in the BIVRTAB = FrequencyCount(A × B) whose size is *TC* × *RC*.
2. Estimate probabilities in the BIVRTAB cells.
3. Compute class-conditional probability $p(r \mid t)$ of reference class $r = 1, ..., RC$, given test class $t = 1, ..., TC$.
4. Reset to zero all $p(r \mid t)$ below $TH1 \in [0, 1]$ (e.g., $TH1 = 9\%$), otherwise set that cell to 1. Let us identify this contingency table instantiation as *DataDrivenConditionalProb*($r|t$)(x, y), x = *1*, ..., *RC*, y = 1, ..., *TC*.
5. Compute class-conditional probability $p(t \mid r)$ of test class $t = 1, ..., TC$, given reference class $r = 1, ..., RC$.
6. Reset to zero all $p(t \mid r)$ below $TH2 \in [0, 1]$ (e.g., $TH2 = 6\% \leq TH1$), otherwise set that cell to 1. Let us identify this contingency table instantiation as *DataDrivenConditionalProb*($t|r$)(x, y), x = *1*, ..., *RC*, y = 1, ..., *TC*.
7. Compute *DataDrivenTemporaryCells*(x, y) = max{*DataDrivenConditionalProb*($t|r$)(x, y), *DataDrivenConditionalProb*($r|t$)(x, y)}, x = *1*, ..., *RC*, y = 1, ..., *TC*. At this point, based exclusively on bottom-up evidence stemming from frequency data, in the 2-fold Cartesian product A × B each cell is equal to 0 or 1. Then that cell is termed either "temporary non-correct" or "temporary correct".
8. Top-down scrutiny by a human domain-expert of each cell in the BIVRTAB, which is either "temporary correct" or "temporary non-correct" at this point, to select those cells to be finally considered as "correct entry-pairs". Actions undertook by this top-down scrutiny are twofold.
   - Switch any data-derived "temporary correct" cell to a "final non-correct" cell if it is provided with a strong prior belief of conceptual mismatch. For example, based on experimental evidence a test spectral category MS white-as-"*Cloud*" can match a reference LC class "*Evergreen Forest*": this data-derived entry-pair match must be considered non-correct in the final R: A ⇒ B following semantic scrutiny by a human expert.
   - Switch any data-derived "temporary non-correct" cell to a "final correct" cell if it is provided with a strong prior belief of conceptual match. For example, the test spectral category MS green-as-"*Vegetation*" is considered a superset of the reference LC class "*Deciduous Forest*" irrespective of whether there are frequency data in support of this conceptual relationship.

Table 1-5 shows an example of how this protocol can be employed in practice. In Table 1-5 the last step 8 identifies an inherently equivocal *information-as-data-interpretation* process, where a human decision maker has a pro-active role in providing frequency data with semantics (meanings) (Capurro and Hjørland 2003). It is highly recommended that any inherently subjective *information-as-data-interpretation* activity occurs as late as possible in the information processing workflow, to avoid propagation of "errors" due to personal preferences not yet community-agreed upon. Noteworthy, in the proposed eight-step guideline there are two "hidden" free-parameters to be user-defined based on heuristics (trial-and-error strategy), the binary thresholds *TH1* and *TH2*, whose normalized range of change and intuitive meaning in terms of probability should make their selection easy and, to a certain extent, application- and user-independent.

## 5. Original measure of association in a categorical variable-pair relationship

Traditional scalar indicators of bivariate categorical variable association estimated from a BIVRTAB = FrequencyCount(A × B), either square or non-square, include the Pearson's chi-square index of statistical independence and the normalized Pearson's chi-square index, also known as Cramer's coefficient V (Sheskin 2000). These frequentist statistics do not apply to a binary relationship R: A ⇒ B such as that shown in Table 1-3. Hereafter, a scalar indicator of association estimated from a binary relationship R: A ⇒ B ⊆ A × B, where A and B are two univariate categorical variables of a single population, is called Categorical Variable Pair Association Index (CVPAI) in range [0, 1].

Proposed in (Baraldi et al. 2014), a CVPAI version 1 (CVPAI1) ∈ [0, 1] is maximized (tends to its best value, equal to 1) if the binary relationship R: A ⇒ B is an injective function (1-1), where test-to-reference class relations as well as reference-to-test class relations are 1-1. An original CVPAI version 2 (CVPAI2) is proposed hereafter as the third original and analytic contribution of the present Part 1. Unlike the CVPAI1, a novel CVPAI2 was constrained as follows, see Figure 1-16. (i) The "most discriminative" test-to-reference class relation is 1-1, e.g., one test color name matches with only one reference LC class name. It means that relationship R: A ⇒ B is a function. (ii) The "most discriminative" reference-to-test class relation is either one-to-many, e.g., one reference LC class matches with several test color names, or 1-1 as a special case of the former, see Figure 1-16. Hence, the CVPAI2 design relaxes the CVPAI1 formulation, i.e., it is always true that CVPAI2 ≥ CVPAI1 ∈ [0, 1]. Whereas the CVPAI1 and CVPAI2 are maximized by different distributions of "correct" entry-pair cells in a binary relationship R: A ⇒ B ⊆ A × B, they are both independent of frequency counts generated by a bivariate categorical variable distribution to be displayed in a BIVRTAB = FrequencyCount(A × B).

To appreciate the conceptual difference between the CVPAI1 and CVPAI2 designs, see Figure 1-16, let us compare a test dictionary A of color names, such as SIAM's, see Figure 1-5, with a reference dictionary B of LC class names, such as NLCD's, see Table 1-2. In terms of capability of color names to discriminate LC class names, the ideal test-to-reference class relation is 1-1, where one color name matches with only one reference LC class. On the other hand, the color attribute of a real-world LC class can be typically linked to one or more discrete color names, see Table 1-3. In this realistic example the expected CVPAI2 value would belong to range (0, 1], while the CVPAI1 formulation proposed in (Baraldi et al. 2014) scores below its maximum, i.e., CVPAI1 ∈ (0, 1).

Another example where the difference between the CVPAI1 and CVPAI2 formulations is highlighted is when the test dictionary A is a specialized version of the reference dictionary B. For example, a test taxonomy of LC classes is A = LegendOfObjectClassNamesA = {LC class "*Dark-tone bare soil"*, LC class "*Light-tone bare soil"*, LC class "*Deciduous Forest"*, LC class "*Evergreen Forest"*} and a reference LC class taxonomy is B = LegendOfObjectClassNamesB = {LC class "*Bare soil"*, LC class "*Forest"*}. Based on our prior knowledge-based understanding of these two semantic dictionaries A and B, a reasonable binary relationship can be considered R: A ⇒ B = {(LC class "*Dark-tone bare soil*", LC class "*Bare soil*"; LC class "*Light-tone bare soil*", LC class "*Bare soil*"; LC class "*Deciduous Forest*", LC class "*Forest*"; LC class "*Evergreen Forest*", LC class "*Forest*")}. In this case, the CVPAI1 formulation scores below its maximum, i.e., CVPAI1 ∈ (0, 1), while the expected CVPAI2 value would score maximum, i.e., CVPAI2 = 1.

These two examples illustrate the intuitive meaning and practical use of the normalized quantitative indicator CVPAI2 ∈ [0, 1] in a hierarchical EO-IUS based on a convergence-of-evidence approach in agreement with Equation (1-3). When the semantic information gap from sub-symbolic sensory data to a symbolic set B = LegendOfObjectClassNames is filled by an EO-IUS starting from a static color naming first stage, provided with a semi-symbolic set A = DictionaryOfColorNames, if the binary relationship R: A ⇒ B features a degree of association CVPAI2 ∈ [0, 1], then (1 - CVPAI2) ∈ [0, 1] is the semantic information gap from sub-symbolic sensory data to the symbolic LegendOfObjectClassNames left to be filled by further stages in the EO-IUS pipeline. If CVPAI2 = 1, then secondary color information discretized by set A = DictionaryOfColorNames suffices to detect the target set B = LegendOfObjectClassNames with no further need to investigate primary spatial information in a convergence-of-evidence image classification approach.

The analytic formulation of the CVPAI2 is proposed as follows. In a binary relationship R: A ⇒ B ⊆ A × B, set A is a test codebook of cardinality |A| = *TC* as rows and set B is a reference codebook of cardinality |B| = *RC* as columns, so that the size of the 2-fold Cartesian product A × B is *TC*×*RC*. The total number of "correct" entry-pair cells in R: A ⇒ B is identified as *CE*, where $0 \le CE \le TC \times RC$. In addition, symbol '==' is adopted to mean 'equal to'. The CVPAI2 formulation is constrained as follows.

(a) $$CE = \sum_{t=1}^{TC} \sum_{r=1}^{RC} CE_{t,r}, \text{with } CE_{t,r} \in \{0,1\} = \{\text{"non-correct" entry-pair } (t,r) = 0, \text{ "correct" entry-pair } (t,r) = 1\},$$
$$CE \in \{0, RC \times TC\}.$$

(b) If (*CE* == 0) then *CVPAI2* = 0 must hold. It means that, when no "correct" cell exists, then the degree of conceptual match between the two categorical variables is zero.

(c) If (*CE* == *TC*×*RC*) then *CVPAI2* → 0 must hold. It means that when all table cells are considered "correct", then no entry-pair is discriminative (informative), i.e., nothing makes the difference between the two categorical variables.

(d) If

$$\left\{\begin{array}{l}\left[\left[\left(\sum_{t=1}^{TC} CE_{t,r} = CE_{+,r}\right) > 0,\ r = 1,\ldots,RC\right] AND\right.\\ \left.\left[\left(\sum_{r=1}^{RC} CE_{t,r} = CE_{t,+}\right) == 1, t = 1,\ldots,TC\right]\right]\end{array}\right\},$$

i.e., for each reference class r = 1, …, RC there is at least one match while each test class t = 1, …, TC features one single match, then *CVPAI2* must be maximum, i.e., *CVPAI2* = 1.

(e) If [*not* condition(b) *AND not* condition(c) *AND not* condition(d)] then *CVPAI2* ∈ (0,1).

In a square binary relationship R: A ⇒ B where TC == RC, to maximize the CVPAI2 (to become equal to 1), submitted to condition (d), the binary relationship must be a 1-1 function (an injective function, 1-1 forward and 1-1 backward). To satisfy the set of aforementioned constraints (a) to (e), the following set of original equations is proposed.

$$CVPSI2 \in [0,1],\ CVPSI2 = \frac{1}{RC+TC}\left(\sum_{r=1}^{RC} f_{RC}(CE_{+,r}) + \sum_{t=1}^{TC} f_{TC}(CE_{t,+})\right), \tag{1-4}$$

with

$$f_{RC}(i) = \begin{cases} 0 \ if \ i=0, \\ 1 \ if \ i>0, \end{cases} \quad i \in \{0,TC\} \subset \mathrm{I}_0^+,\ \text{where } i = CE_{+,r}, r \in \{1,RC\}, \tag{1-5}$$

$$f_{TC}(j) = \begin{cases} 0 \ if \ j = 0, \\ GaussianMembership(j, Center = 1, StnDev = RC/3) = \\ e^{-\frac{1}{2}\frac{(j-1)^2}{\left(\frac{RC}{3}\right)^2}} \in [0,1], \quad if \ j > 0, \text{with } j \in \{0, RC\} \subset \mathrm{I}_0^+, \text{where } j = CE_{t,+}, t \in \{1, TC\}.. \end{cases} \tag{1-6}$$

In Equation (1-6) GaussianMembership(j, Center = 1, StnDev = RC/3) → 0 when it covers approximately 99.73% of the area underneath the curve at distance j ≈ ± 3 • StnDev = ± 3 • RC / 3 = ± RC from its Center = 1. If j = 1, then GaussianMembership(j, Center = 1, StnDev = RC/3) = 1. It is trivial to prove that Equation (1-4) to Equation (1-6) satisfy the aforementioned requirements (a) to (d). By means of a numeric example, it can be shown that requirement (e) is satisfied too. For example, estimated from the binary relationship instantiated at step 8 of Table 1-5, CVPAI2 = (1/6)*(1 + 1 + 1 + 1 + 1 + exp(-0.5*(3-1)^2/(3/3)^2)) = (1/6) * (5 + 0.1353) = 0.8558. Intuitively closer to 1 than 0 this CVPAI2 value shows that the harmonization between the two test and reference nominal variables is (fuzzy) "high" (> 0.8) in Table 1-5.

## 6. Conclusions

To pursue a GEOSS mission not-yet accomplished by the RS community, this interdisciplinary work aimed at filling an analytic and pragmatic information gap from EO big sensory data to the ESA EO Level 2 product. For the sake of readability this paper is split into two, the present Part 1 – Theory and the following Part - Validation.

The original contribution of the present Part 1 is fourfold. A first lesson was learned from published works on prior knowledge-based MS reflectance space hyperpolyhedralization into static (non-adaptive-to-data) color names. It was observed that well-established RS practices, such as 1D image analysis based on supervised data learning algorithms, where dominant spatial information is neglected in favor of secondary color information, and thematic map quality assessment where test and reference map legends are required to coincide, can become malpractices when an *a priori* dictionary of static color names is employed for MS image classification purposes in agreement with Equation (1-3). When test and reference thematic map legends A and B are the same, the binary relationship R: A $\Rightarrow$ B $\subseteq$ A $\times$ B becomes a bijective function (both 1-1 and onto) and the main diagonal of the 2-fold Cartesian product A $\times$ B guides the interpretation process of a BIVRTAB = FrequencyCount(A $\times$ B) = CMTRX. This constraint makes a CMTRX intuitive to understand and more familiar to RS practitioners. Quite surprisingly, the non-coincident assumption between an *a priori* dictionary A of static sub-symbolic color names and a scene- and application-specific legend B of symbolic classes of real-world objects appears somehow difficult to acknowledge by relevant portions of the RS community, in contrast with common sense, see Table 1-3.

Second, Equation (1-3) was proposed as an analytic expression of a biologically plausible hybrid (combine deductive and inductive) CV system suitable for convergence of color and spatial evidence, in agreement with the statistic stratification principle and the divide-and-conquer problem solving approach. In compliance with common sense (see Table 1-3), Equation (1-3) shows that a static color naming first stage can be employed for stratification purposes of further spatial-context sensitive image classification stages. In the static color naming first stage, a binary relationship R: A $\Rightarrow$ B $\subseteq$ A $\times$ B from a set A of general-purpose static color names to a set B of user- and application-specific LC class names can be established by human experts based on top-down prior beliefs, if any, in combination with bottom-up evidence inferred from new data.

Third, for best practice an eight-step protocol was streamlined to infer a categorical variable-pair relationship R: A $\Rightarrow$ B from categorical variable A to categorical variable B as a hybrid combination of deductive prior beliefs with inductive evidence from data.

Fourth, an original CVPAI2 formulation was proposed as a categorical variable-pair degree of association in a binary relationship R: A $\Rightarrow$ B.

To comply with the QA4EO *Cal*/*Val* requirements, the subsequent Part 2 of this paper presents and discusses a Stage 4 *Val* of the SIAM-WELD annual map time-series in comparison with the reference NLCD 2006 map, based on an original protocol for wall-to-wall inter-map comparison without sampling where the test and reference maps feature the same spatial resolution and spatial extent, but whose legends are not the same and must be harmonized.

## Acknowledgments

To accomplish this work Andrea Baraldi was supported in part by the National Aeronautics and Space Administration (NASA) under Grant No. NNX07AV19G issued through the Earth Science Division of the Science Mission Directorate. Dirk Tiede was supported in part by the Austrian Research Promotion Agency (FFG), in the frame of project AutoSentinel2/3, ID 848009. Prof. Ralph Maughan, Idaho State University, is kindly acknowledged for his contribution as active conservationist and for his willingness to share his invaluable photo archive with the scientific community as well as the general public. Andrea Baraldi thanks Prof. Raphael Capurro, Hochschule der Medien, Germany, and Prof. Christopher Justice, Chair of the Department of Geographical Sciences, University of Maryland, for their support. Above all, the authors acknowledge the fundamental contribution of Prof. Luigi Boschetti, currently at the Department of Forest, Rangeland and Fire Sciences, University of Idaho, Moscow, Idaho, who conducted by independent means all experiments whose results are proposed in this validation paper. The authors also wish to thank the Editor-in-Chief, Associate Editor and reviewers for their competence, patience and willingness to help.

## Disclosure statement

In accordance with XXX policy and his ethical obligation as a researcher, Andrea Baraldi reports he is the sole developer and IPR owner of the Satellite Image Automatic Mapper™ (non-registered trademark) computer program licensed to academia, public institutions and private companies, eventually free-of-charge, by the one-man-company Baraldi Consultancy in Remote Sensing that may be affected by the research reported in the enclosed paper. Andrea Baraldi has disclosed those interests fully to XXX, and he has in place an approved plan for managing any potential conflicts arising from that involvement.

## Figures and figure captions

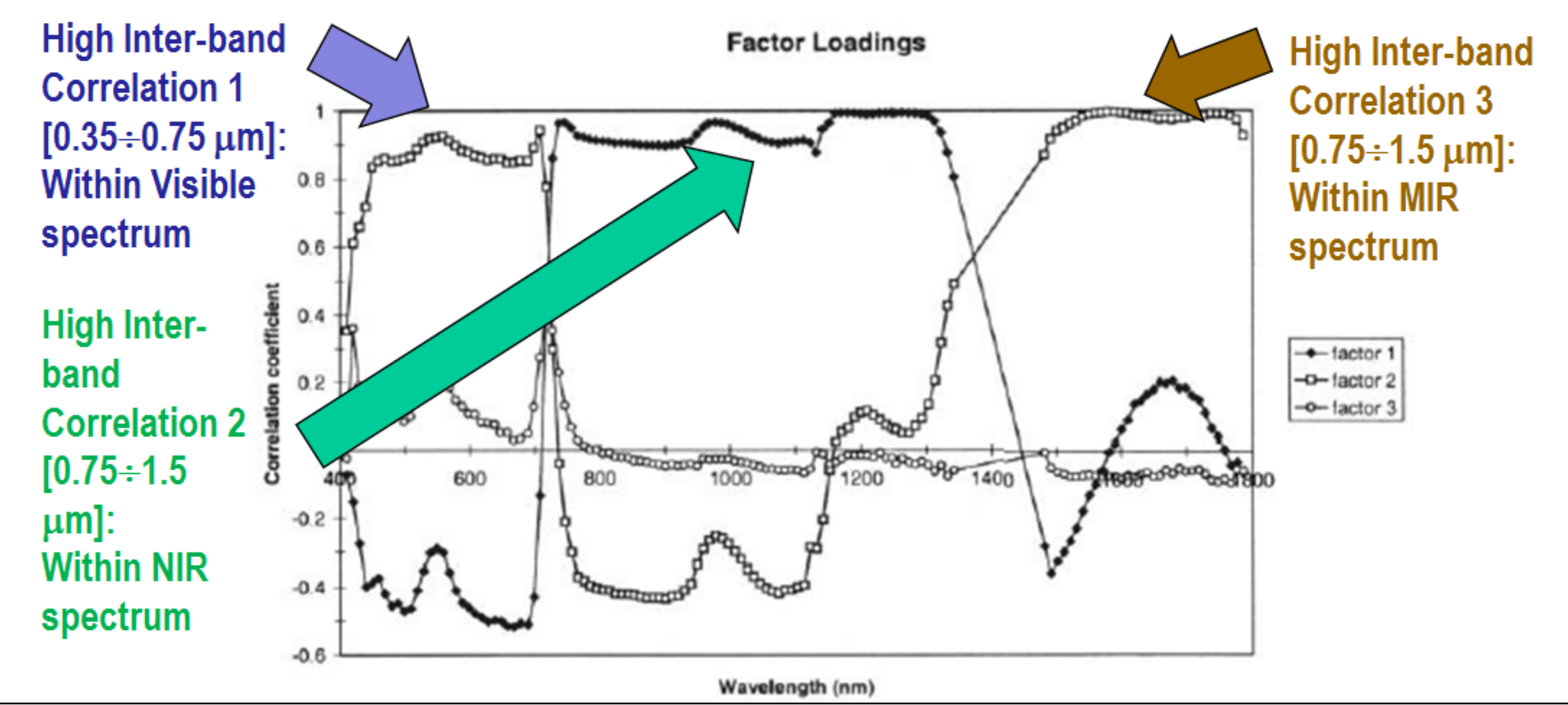

Figure 1-1. Courtesy of van der Meer and De John (2000). Pearson's cross-correlation (CC) coefficients for the main factors resulting from a principal component analysis and factor rotation for an agricultural data set based on spectral bands of the AVIRIS hyper-spectral (HS) spectrometers 1, 2 and 3. Flevoland test site, July 5th 1991. Inter-band CC values are "high" (> 0.8) within the visible spectral range, the Near Infra-Red (NIR) wavelenghts and the Medium IR (MIR) wavelengths. The general conclusion is that, irrespective of non-stationary local information, the global (image-wide) information content of a multi-spectral (MS) image whose number $N$ of spectral channels $\in \{2, 9\}$, a super-spectral (SS) image with $N \in \{10, 20\}$, or an hyperspectral (HS) image with $N > 20$, can be preserved by selecting one visible, one NIR, one MIR and one thermal IR (TIR) band, such as in the spectral resolution of the National Oceanic and Atmospheric Administration (NOAA) Advanced Very High Resolution Radiometer (AVHRR) imaging sensor series in operating mode from 1978 to date.

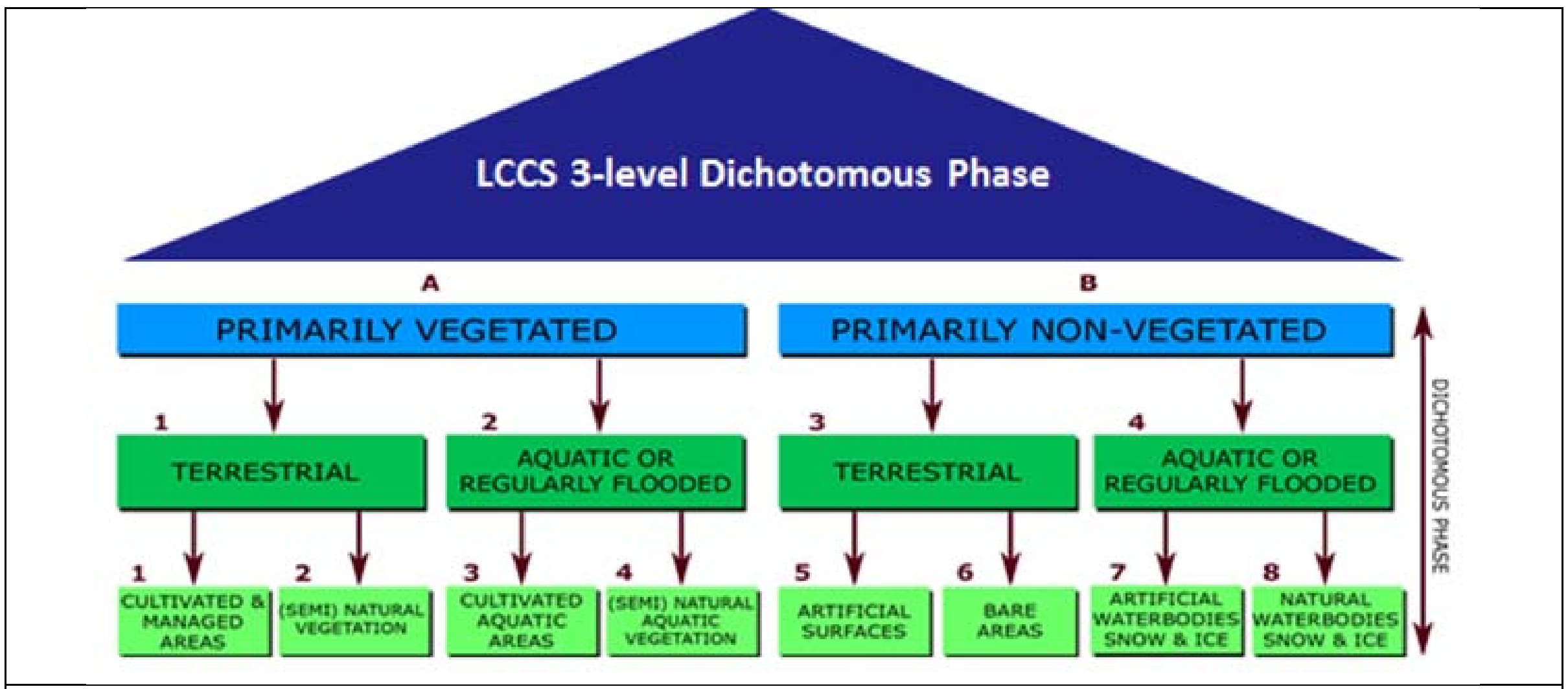

Figure 1-2. The nested 3-level LCCS-DP layers are: (i) vegetation versus non-vegetation, (ii) terrestrial versus aquatic, and (iii) managed versus natural or semi-natural. They deliver as output the following 8-class LCCS-DP taxonomy. (A11) Cultivated and Managed Terrestrial (non-aquatic) Vegetated Areas. (A12) Natural and Semi-Natural Terrestrial Vegetation. (A23) Cultivated Aquatic or Regularly Flooded Vegetated Areas. (A24) Natural and Semi-Natural Aquatic or Regularly Flooded Vegetation. (B35) Artificial Surfaces and Associated Areas. (B36) Bare Areas. (B47) Artificial Waterbodies, Snow and Ice. (B48) Natural Waterbodies, Snow and Ice. The general-

purpose user- and application-independent 8-class LCCS-DP taxonomy is preliminary to a user- and application-specific LCCS Modular Hierarchical Phase (MHP) taxonomy.

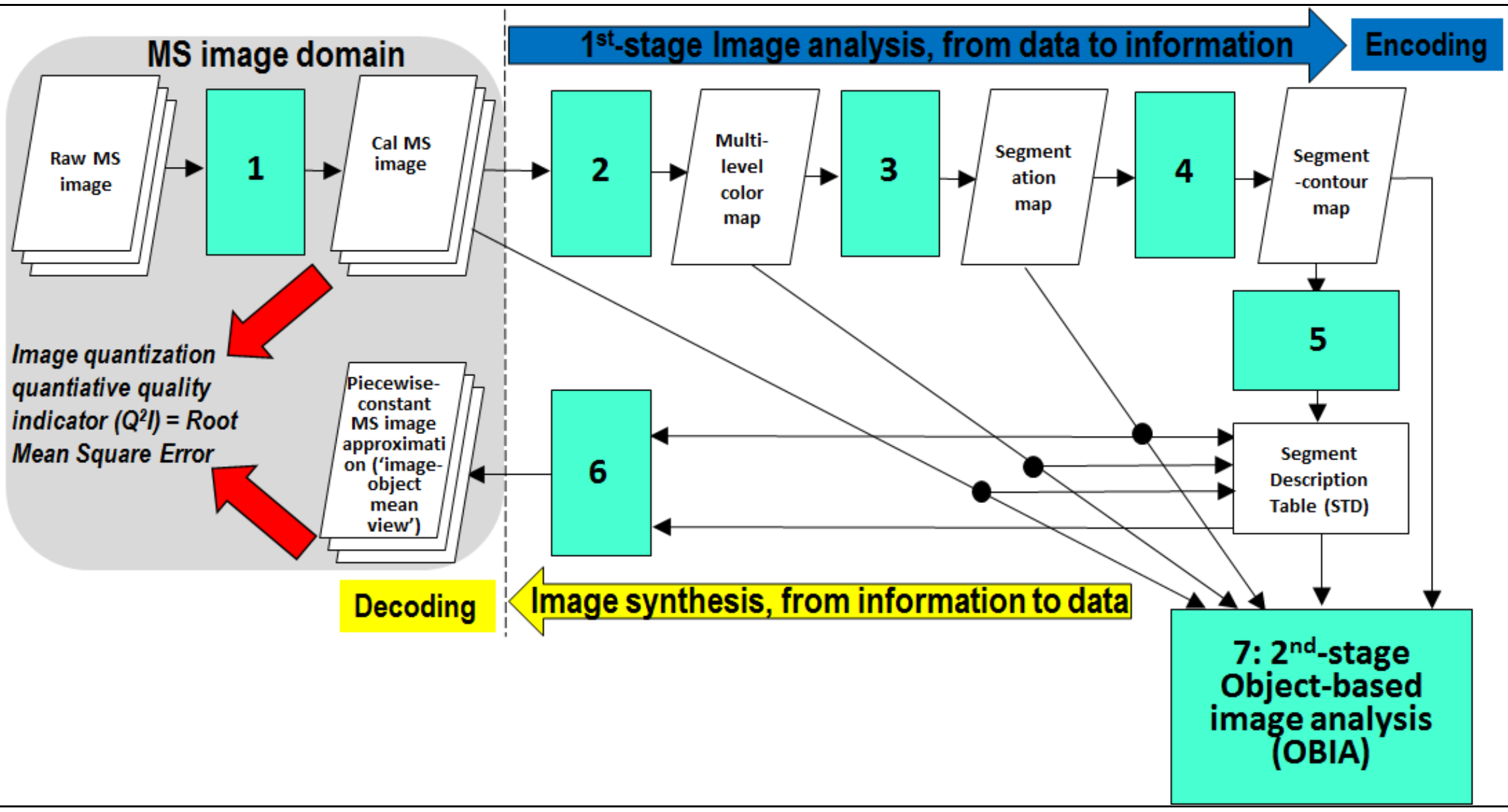

Figure 1-3. The SIAM application software for prior knowledge-based vector quantization (VQ) in a radiometrically calibrated MS data space. It consists of six subsystems, identified as 1 to 6. Phase 1-of-2 = Encoding phase/Image analysis - Stage 1: MS data calibration into top-of-atmosphere reflectance (TOARF) or surface reflectance (SURF) values. Stage 2: Prior knowledge-based SIAM decision tree for MS reflectance space partitioning (quantization, hyperpolyhedralization). Stage 3: Well-posed (deterministic) two-pass connected-component detection in the multi-level color map-domain. Connected-components in the color map-domain are connected sets of pixels featuring the same color label. These connected-components are also called image-objects, segments or superpixels. Stage 4: Well-posed superpixel-contour extraction. Stage 5: Superpixel description table allocation and initialization. Phase 2-of-2 = Decoding phase/Image synthesis - Stage 6: Superpixelwise-constant input image approximation ("image-object mean view') and per-pixel VQ error estimation. (Stage 7: in cascade to the SIAM superpixel detection, a high-level object-based image analysis (OBIA) approach can be adopted).

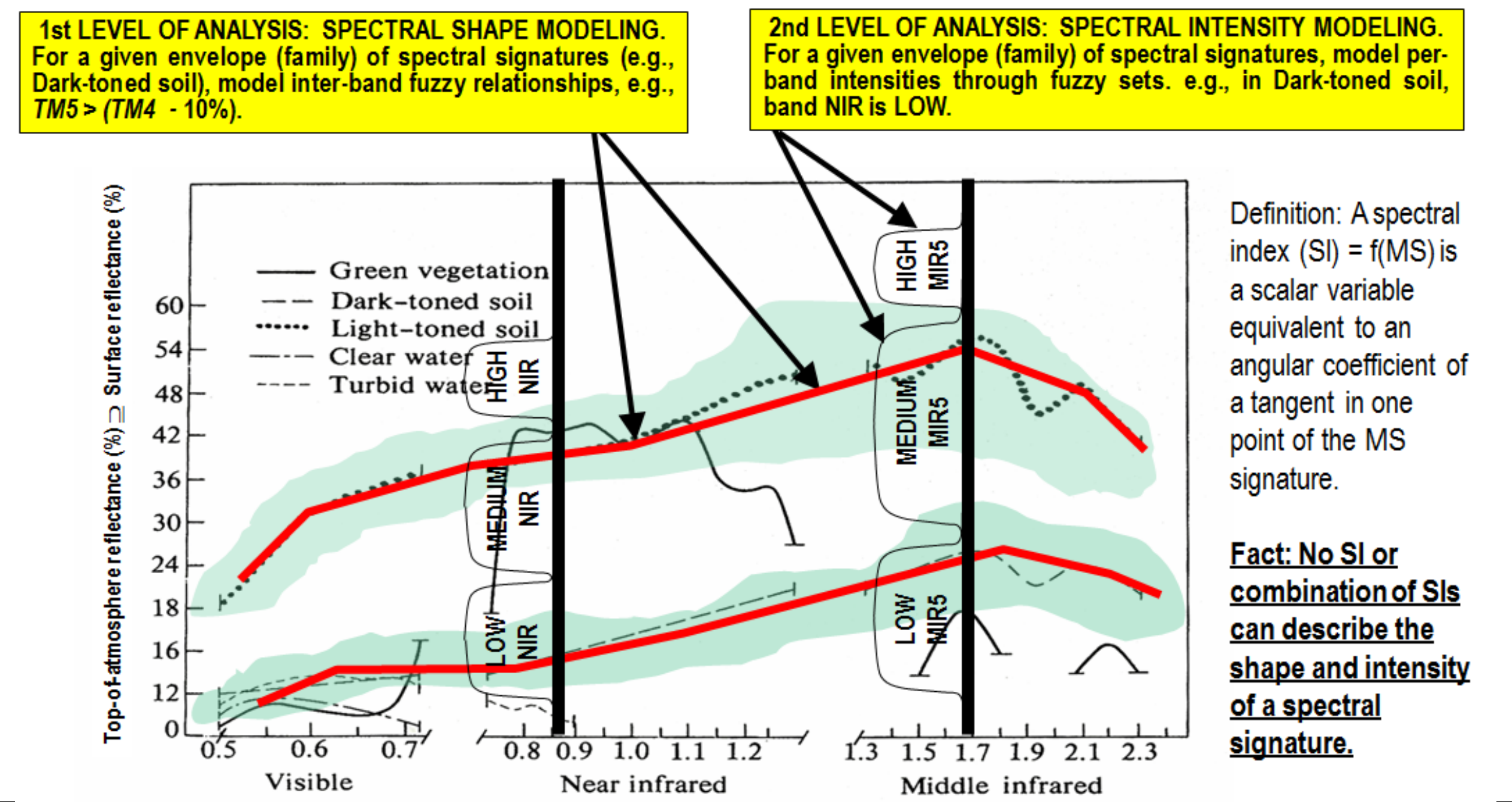

Figure 1-4. Examples of land cover (LC)-class specific families of spectral signatures in top-of-atmosphere reflectance (TOARF) values which include surface reflectance (SURF) values as a special case in clear sky and flat terrain conditions. A within-class family of spectral signatures (e.g., dark-toned soil) in TOARF values forms a buffer zone (hyperpolyhedron, envelope, manifold). The SIAM decision tree models each target family of spectral signatures in terms of multivariate shape and multivariate intensity as a viable alternative to multivariate analysis of spectral indexes. A typical spectral index is a scalar band ratio equivalent to an angular coefficient of a tangent in one point of the spectral signature. Infinite functions can feature the same tangent value in one point. In practice, no spectral index or combination of spectral indexes can reconstruct the multivariate shape and multivariate intensity of a spectral signature.

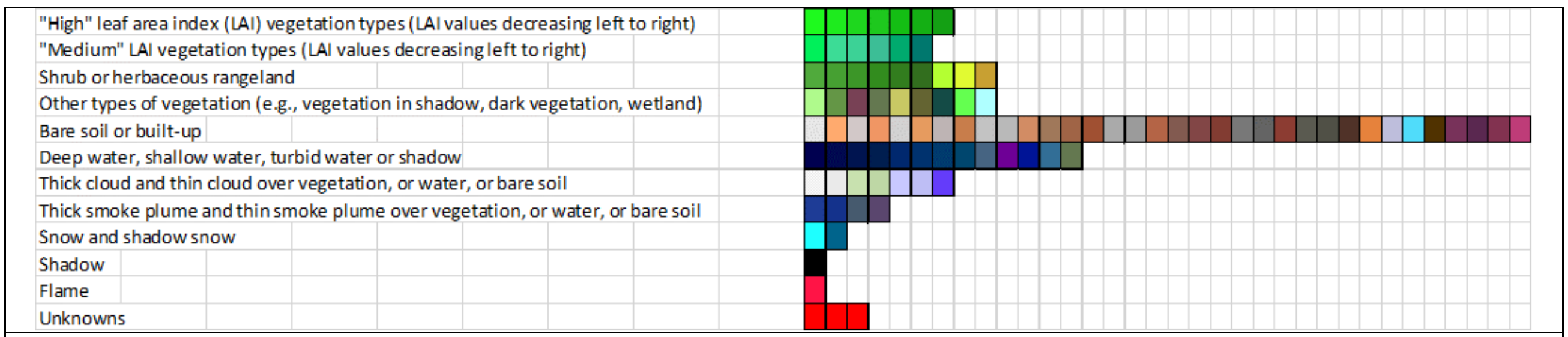

Figure 1-5. Prior knowledge-based color map legend adopted by the Landsat-like SIAM (L-SIAM™, release 88 version 6) implementation. For the sake of representation compactness pseudo-colors of the 96 spectral categories are gathered along the same raw if they share the same parent spectral category in the decision tree, e.g., "strong" vegetation, equivalent to a spectral end-member. The pseudo-color of a spectral category is chosen as to mimic natural colors of pixels belonging to that spectral category. These 96 color names at fine color granularity are aggregated into 48 and 18 color names at intermediate and coarse color granularity respectively, according to parent-child relationships defined *a priori*, also refer to Table 1-1.

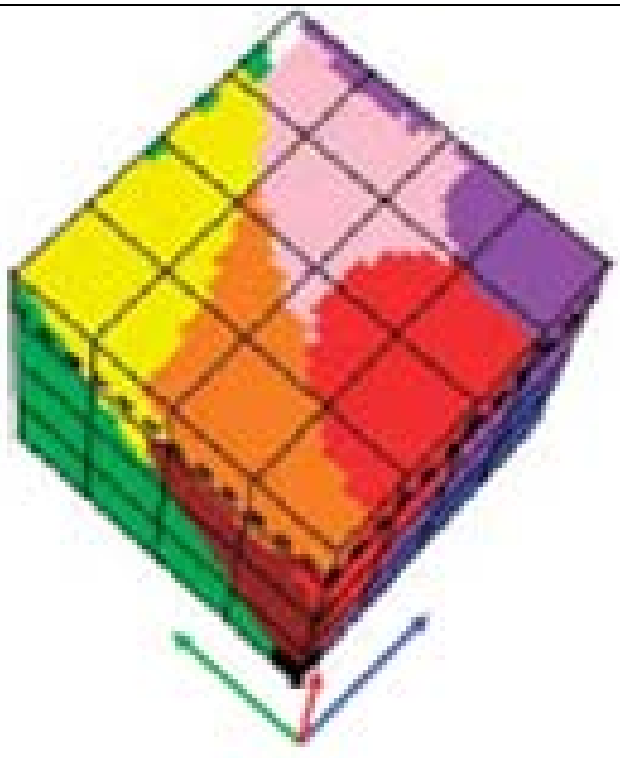

Figure 1-6. Courtesy of Griffin (2006). Monitor-typical RGB cube partitioned into perceptual polyhedra corresponding to a discrete and finite dictionary of basic color (BC) names, to be community-agreed upon in advance to be employed by members of the community. The mutually exclusive and totally exhaustive polyhedra are neither necessarily convex nor connected. In practice BC names belonging to a finite and discrete color dictionary are equivalent to Vector Quantization (VQ) levels belonging to a VQ codebook (Cherkassky and Mulier 1998).

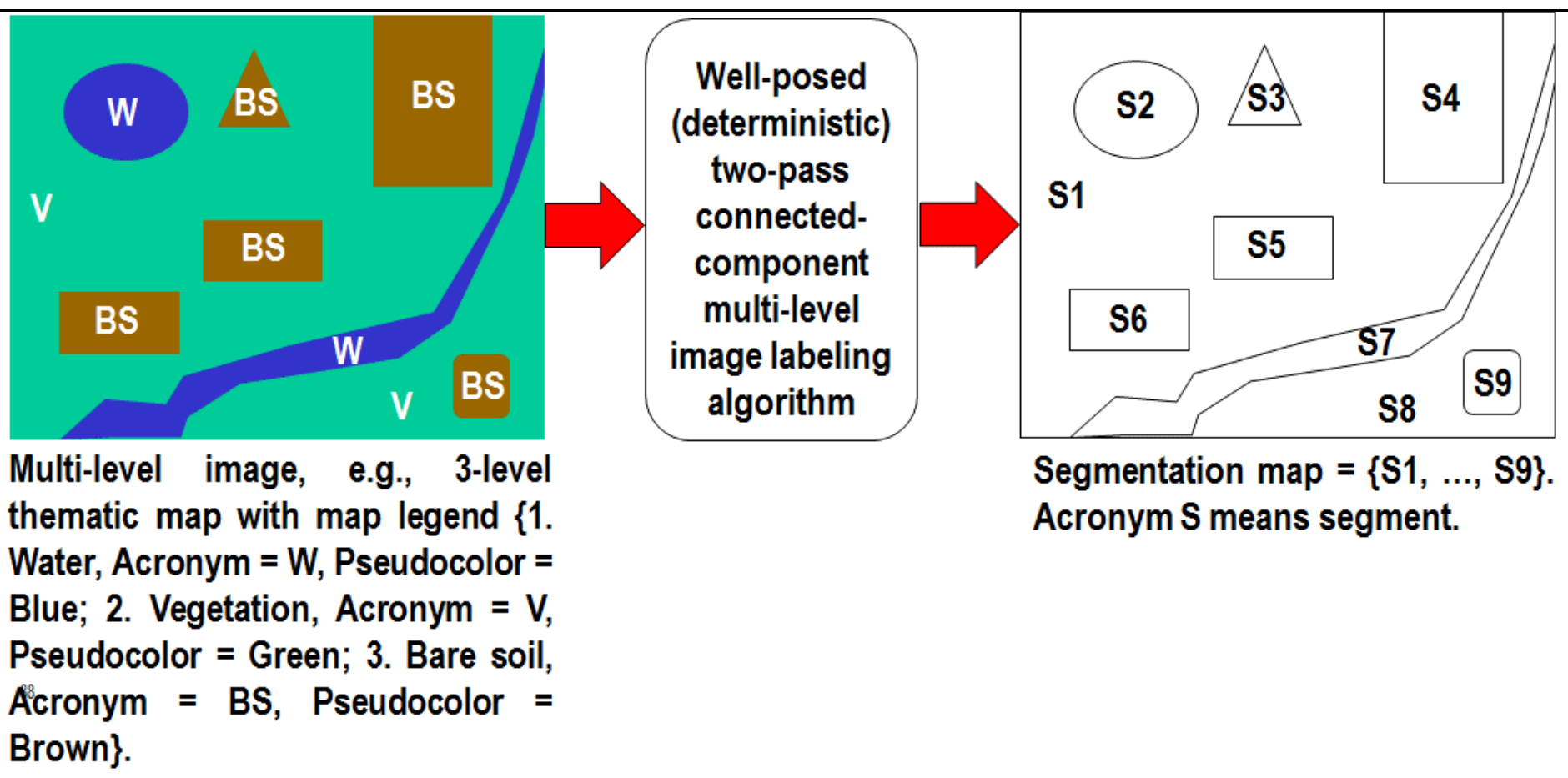

Figure 1-7. One segmentation map is deterministically generated from one multi-level image, such as a thematic map, but the vice versa does not hold, i.e., many multi-level images can generate the same segmentation map. In this example, nine image-objects/segments S1 to S9 can be detected in the 3-level thematic map shown at left. Each segment consists of a connected set of pixels sharing the same multi-level map label. Each stratum/layer/level consists of one or more segments, e.g., stratum Vegetation (V) consists of two disjoint segments, S1 and S8. In any multi-level (categorical, nominal, qualitative) image domain, three labeled spatial primitives co-exist and are provided with parent-child relationships: pixel with a level-label and a pixel identifier (ID, e.g., the row-column coordinate pair), segment (polygon) with a level-label and a segment ID, and stratum (multi-part polygon) with a level-label equivalent to a stratum ID. This overcomes the ill-fated dichotomy between traditional unlabeled sub-symbolic pixels versus labeled sub-symbolic segments in the numeric (quantitative) image domain traditionally coped with by the object-based image analysis (OBIA) paradigm (Blaschke et al. 2014).

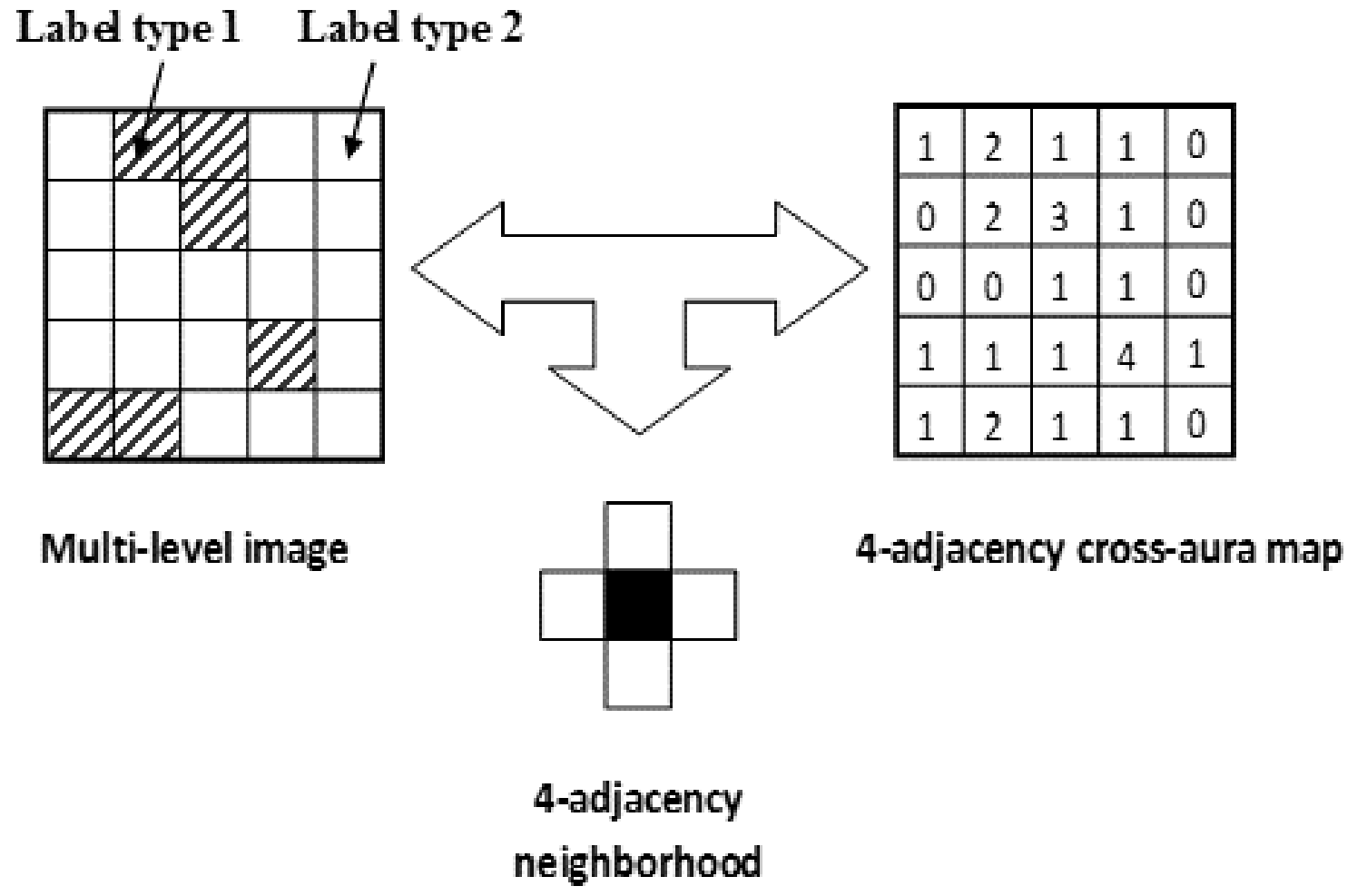

Figure 1-8. Example of a 4-adjacency cross-aura map, shown at right, generated in linear time from a two-level image shown at left.

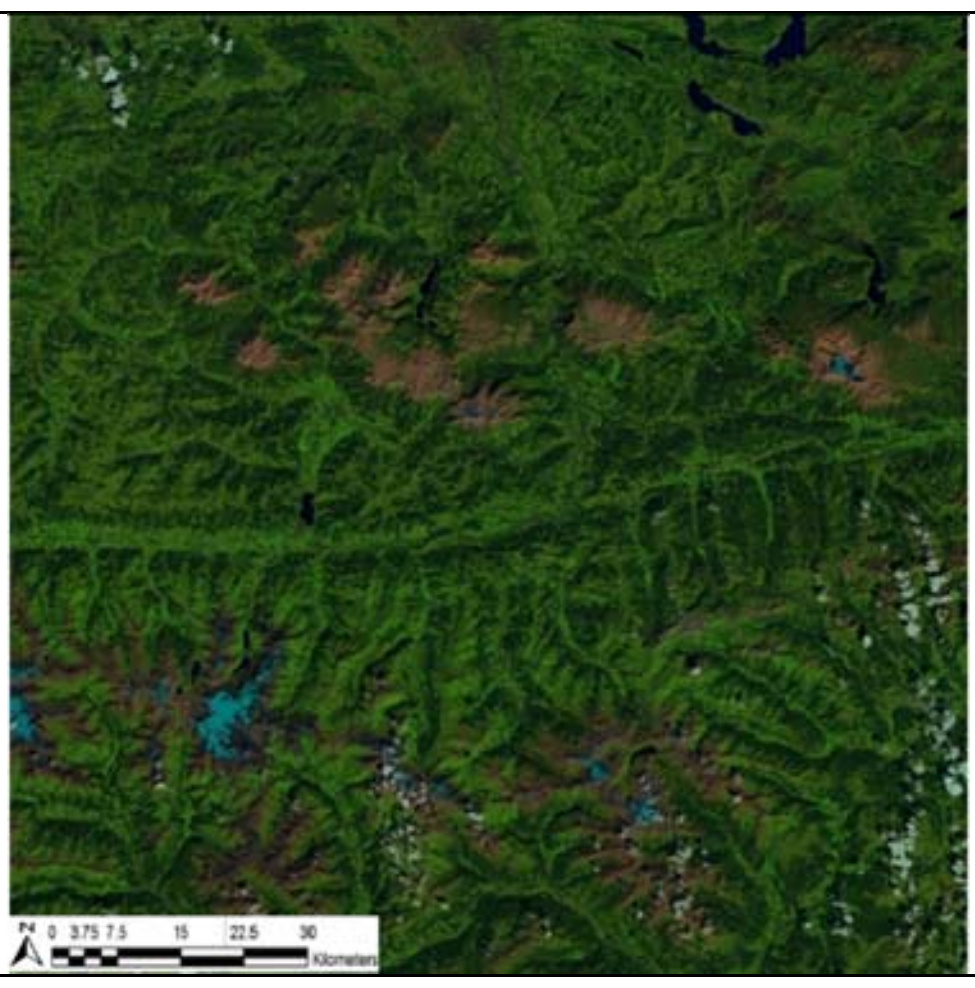

(a)

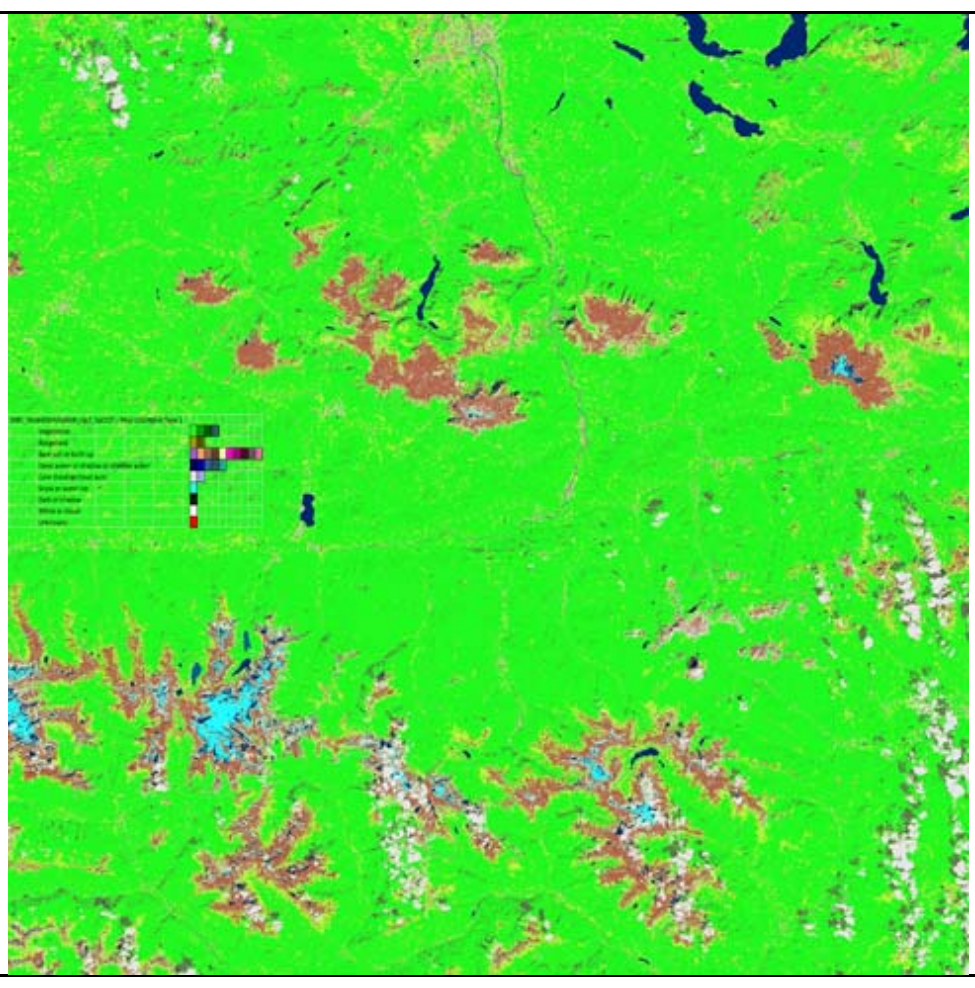

(b)

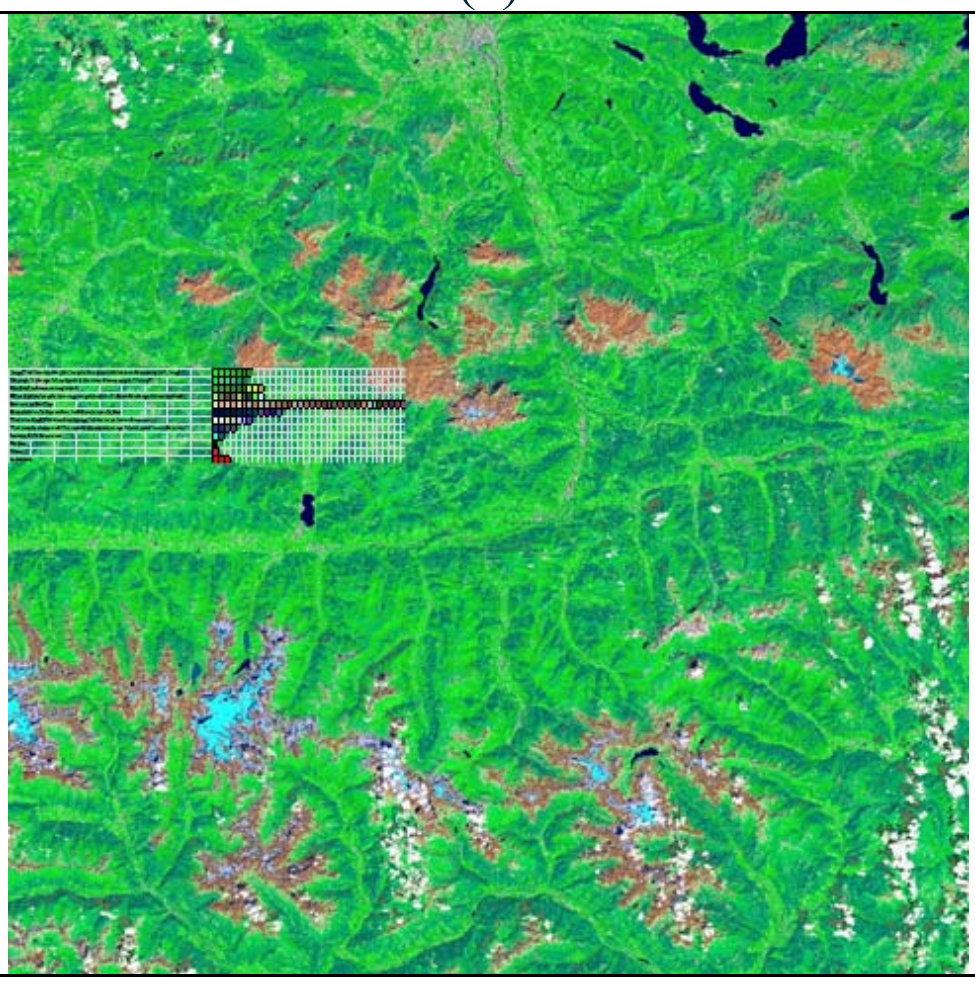

(c)

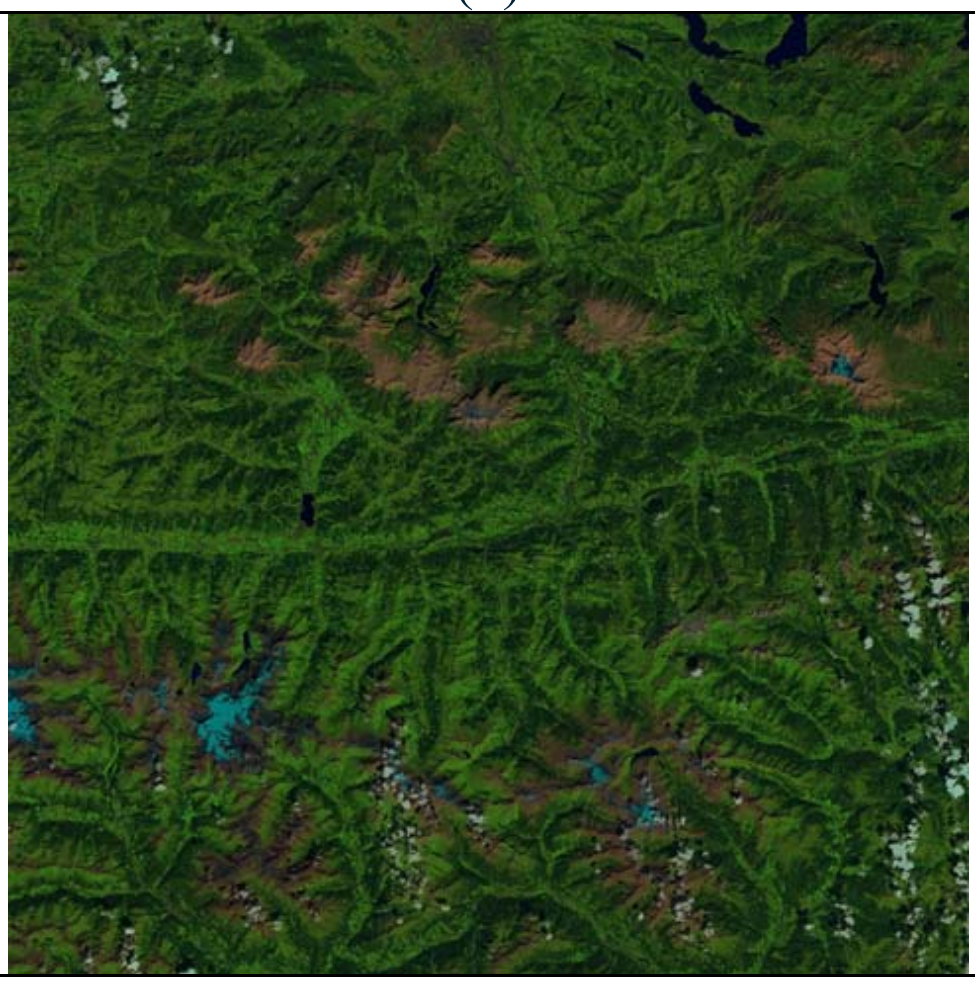

(d)

| | |
|---|---|
| 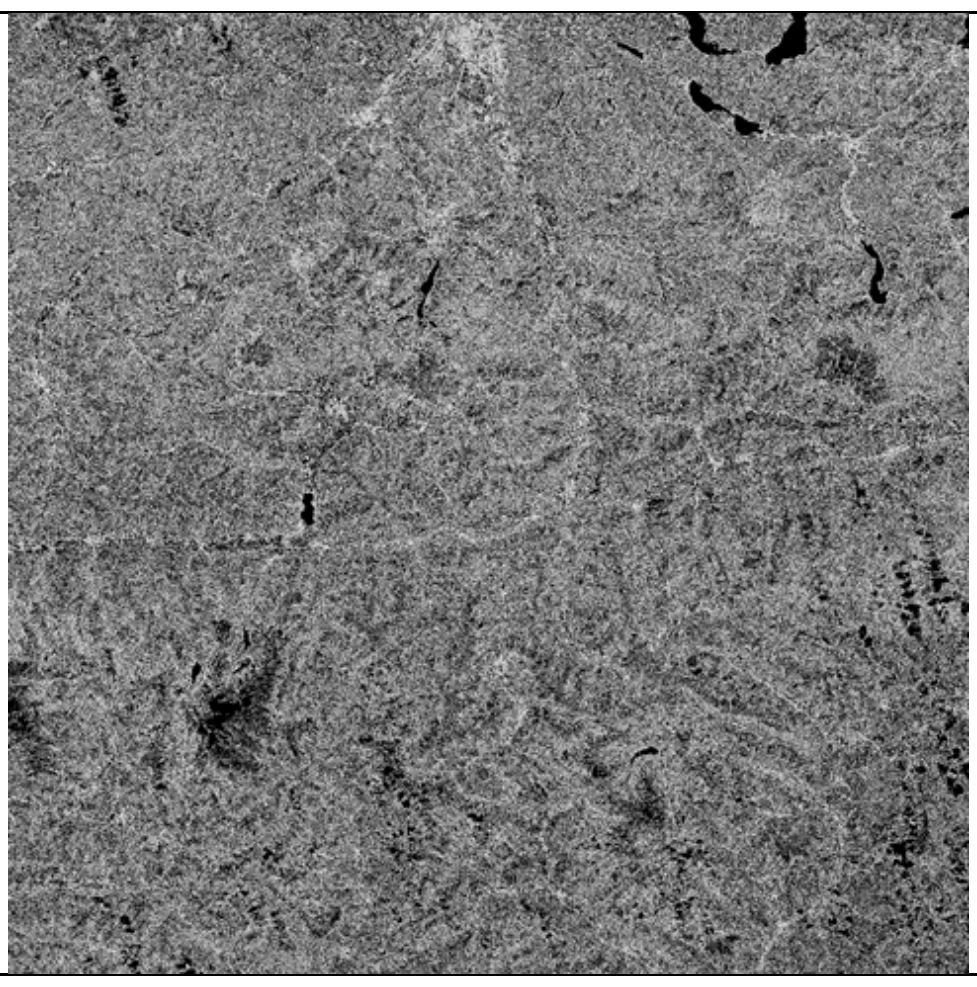 | 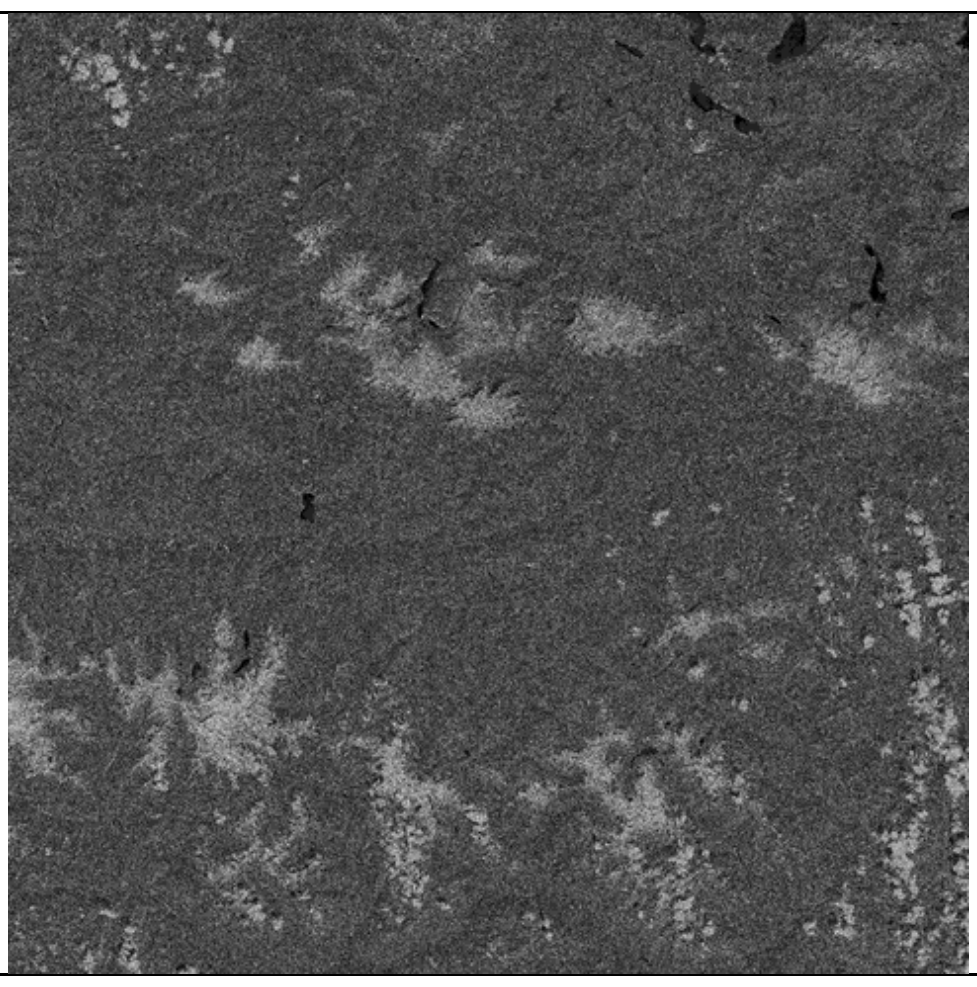 |
| (e) | (f) |

Figure 1-9. (a) Sentinel-2A MSI Level-1C image of the Earth surface south of the city of Salzburg, Austria. The city area is visible around the middle of the image upper boundary (Lat-long coordinates: 47°48'25.0"N 13°02'43.6"E). Acquired on 2015-09-11. Spatial resolution: 10 m. Image size: 110×110 km. Radiometrically calibrated into TOARF values in range {0, 255}, it is depicted as a false color RGB image, where: R = Medium InfraRed (MIR) = Band 11, G = Near IR (NIR) = Band 8, B = Blue = Band 2. No histogram stretching is applied for visualization purposes. (b) L-SIAM color map at coarse color granularity, consisting of 18 spectral categories depicted in pseudo colors shown in the map legend. Coarse-granularity color categories are generated by merging color hyperpolyhedra at fine color granularity, according to pre-defined parent-child relationships, refer to Table 1-1. (c) L-SIAM color map at fine color granularity, consisting of 96 spectral categories depicted in pseudo colors shown in the map legend. (d) Superpixelwise-constant approximation of the input image ("image-object mean view") generated from the L-SIAM's 96 color map at fine granularity. Depicted in false colors: R = MIR = Band 11, G = NIR = Band 8, B = Blue = Band 2. Spatial resolution: 10 m. No histogram stretching is applied for visualization purposes. (e) 8-adjacency cross-aura contour map in range {0, 8} automatically generated from the L-SIAM's 96 color map at fine granularity. It shows contours of connected sets of pixels featuring the same color label. These connected-components are also called image-objects, segments or superpixels. (f) Per-pixel scalar difference between the input MS image shown in (a) and the superpixelwise-constant MS image reconstruction shown in (d). This scalar difference is computed as the per-pixel Root Mean Square Error (RMSE) in range {0, 255}. The RMSE is a well-known vector quantization (VQ) error. Image-wide basic statistics: Min = 0, Max = 130, Mean = 2.60, Stdev = 3.45. Histogram stretching is applied for visualization purposes.

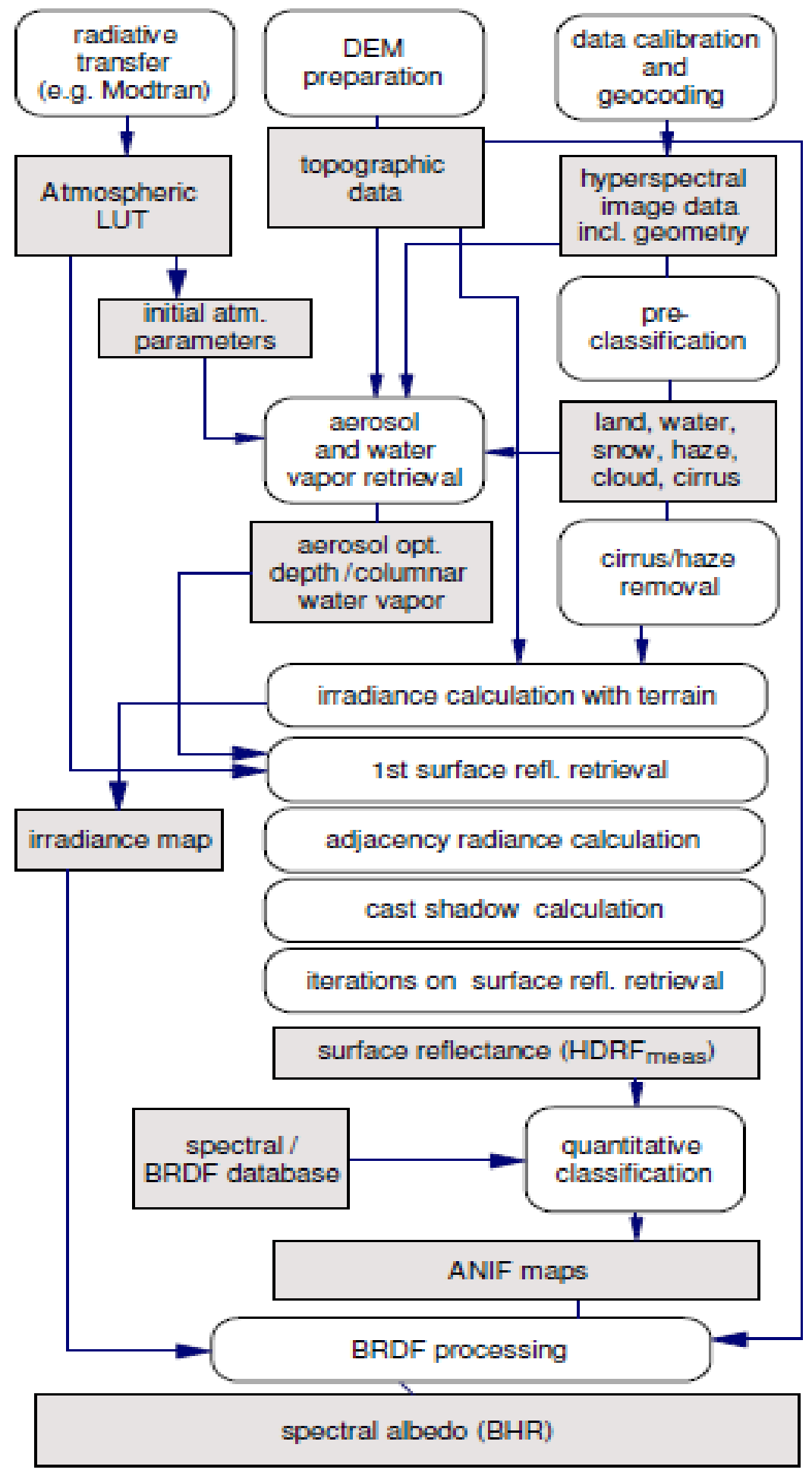

Figure 1-10. Same as in Richter et al. (2009), courtesy of Daniel Schläpfer, ReSe Applications Schläpfer. A complete ("augmented") hybrid inference workflow for MS image correction from atmospheric, adjacency and topographic effects. It combines a standard Atmospheric/Topographic Correction for Satellite Imagery (ATCOR) commercial software workflow (Richter and D. Schläpfer 2012a; Richter and D. Schläpfer 2012b), with a bidirectional reflectance distribution function (BRDF) effect correction. Processing blocks are represented as circles and output products as rectangles. This hybrid (combined deductive and inductive) workflow alternates deductive/prior knowledge-based and inductive/learning-from-data inference units, starting from initial conditions provided by a first-stage deductive Spectral Classification of surface reflectance signatures (SPECL) decision tree for color naming (pre-classification), implemented within the ATCOR commercial software toolbox (Richter and D. Schläpfer 2012a; Richter and D. Schläpfer 2012b). Categorical variables generated by the pre-classification and classification blocks are employed to stratify (mask) unconditional numeric variable distributions, in line with the statistic stratification principle (Hunt and Tyrrell 2012). Through statistic stratification, inherently ill-posed inductive learning-from-data algorithms are provided with prior knowledge required in addition to data to become better posed for numerical solution, in agreement with the machine learning-from-data literature (Cherkassky and Mulier 1998).

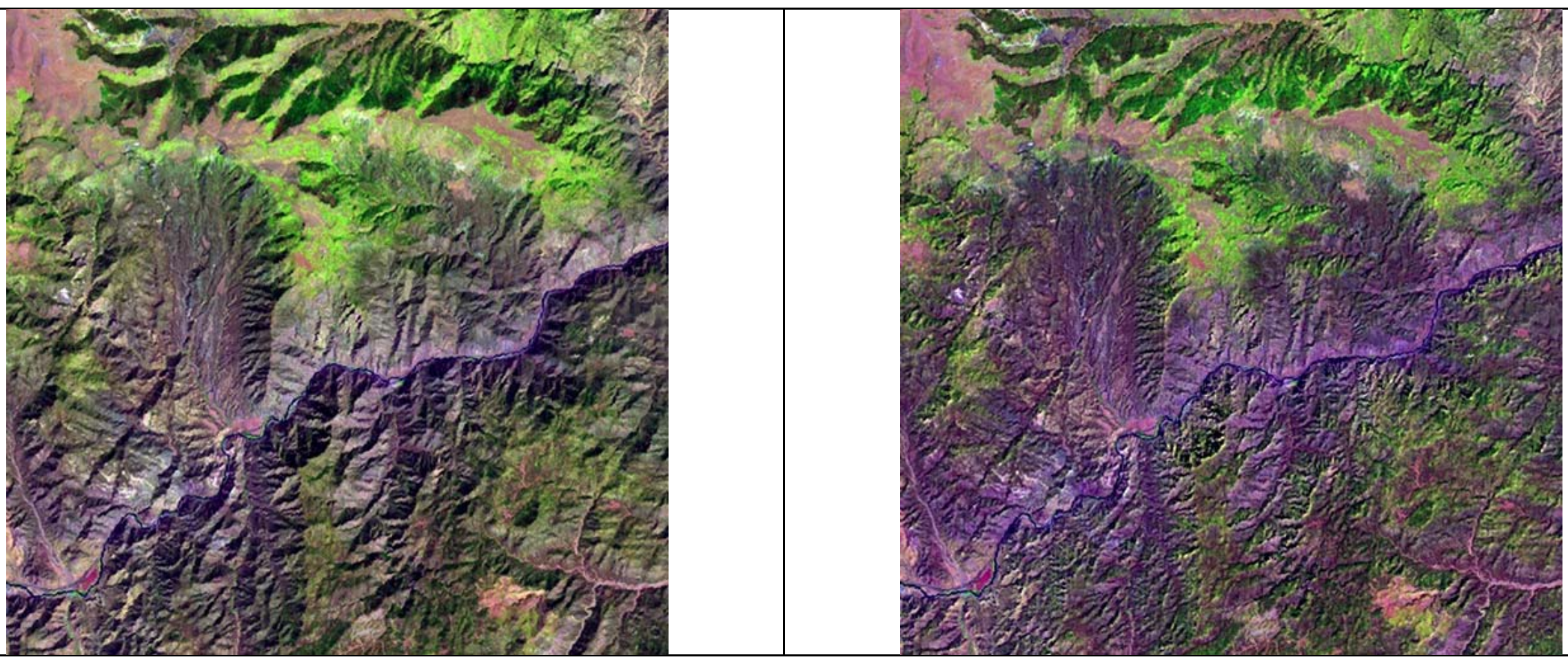

Figure 1-11. Left: Zoomed area of a Landsat 7 ETM+ image of Colorado, USA (path: 128, row: 021, acquisition date: 2000-08-09), depicted in false colors (R: band ETM5, G: band ETM4, B: band ETM1), 30 m resolution, radiometrically calibrated into TOARF values. Right: Output product automatically generated without human-machine interaction by the stratified topographic correction (STOC) algorithm proposed in Baraldi et al. (2010), whose input datasets are one Landsat image, its data-derived L-SIAM color map at coarse color granularity, consisting of 18 spectral categories for stratification purposes (see Table 1-1), and a standard 30 m resolution Shuttle Radar Topography Mission (SRTM) digital elevation model (DEM).

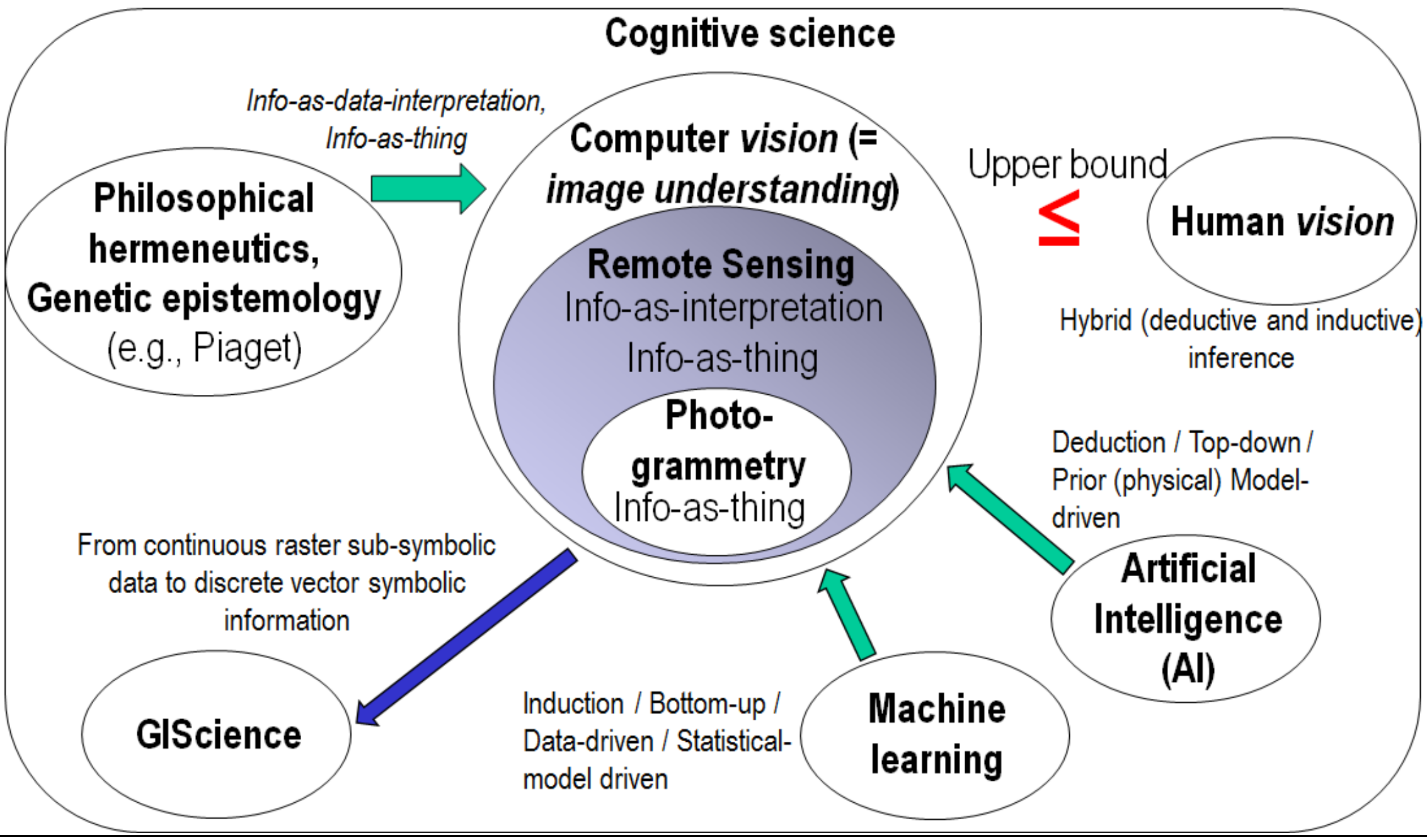

Figure 1-12. Like engineering, RS is a metascience, whose goal is to transform knowledge of the world, provided by other scientific disciplines, into useful user- and context-dependent solutions in the world. Cognitive science is the interdisciplinary scientific study of the mind and its processes. It examines what cognition (learning) is, what it does and how it works. It especially focuses on how information/knowledge is represented, acquired, processed and transferred within nervous systems (distributed processing systems in humans, such as the human brain, or other animals) and machines (e.g., computers). Neurophysiology studies nervous systems, including the brain.

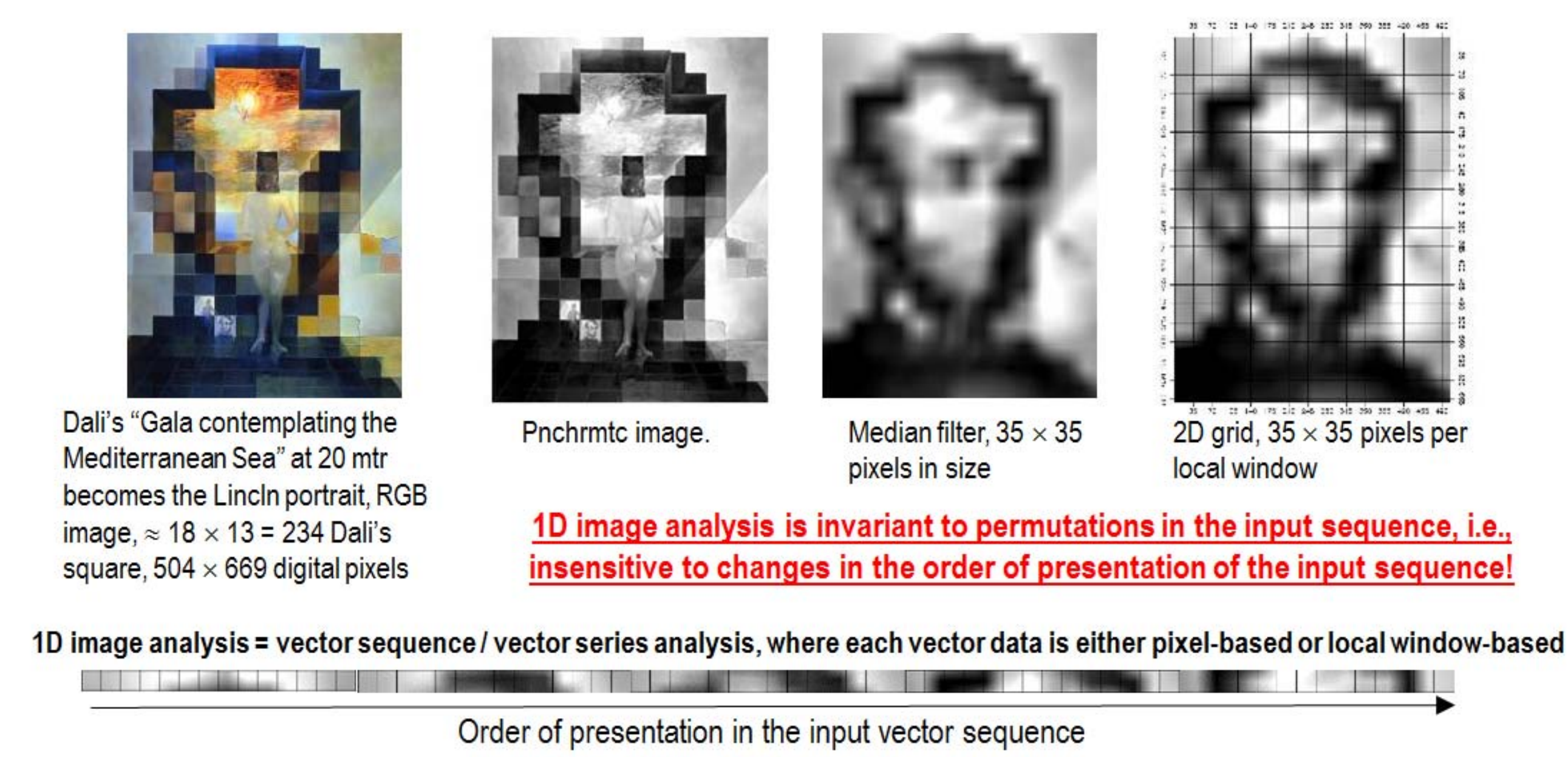

Figure 1-13. Example of 1D image analysis. The (2D) image at left is transformed into the 1D vector data stream shown at bottom, where vector data are either pixel-based or spatial context-sensitive, e.g., local window-based. This 1D vector data stream means nothing to a human photointerpreter. When it is input to a traditional inductive data learning classifier, it is what the inductive classifier actually sees when watching the (2D) image at left. Undoubtedly, computers are more successful than humans in 1D image analysis. Nonetheless, humans are still far more successful than computers in (2D) image analysis.

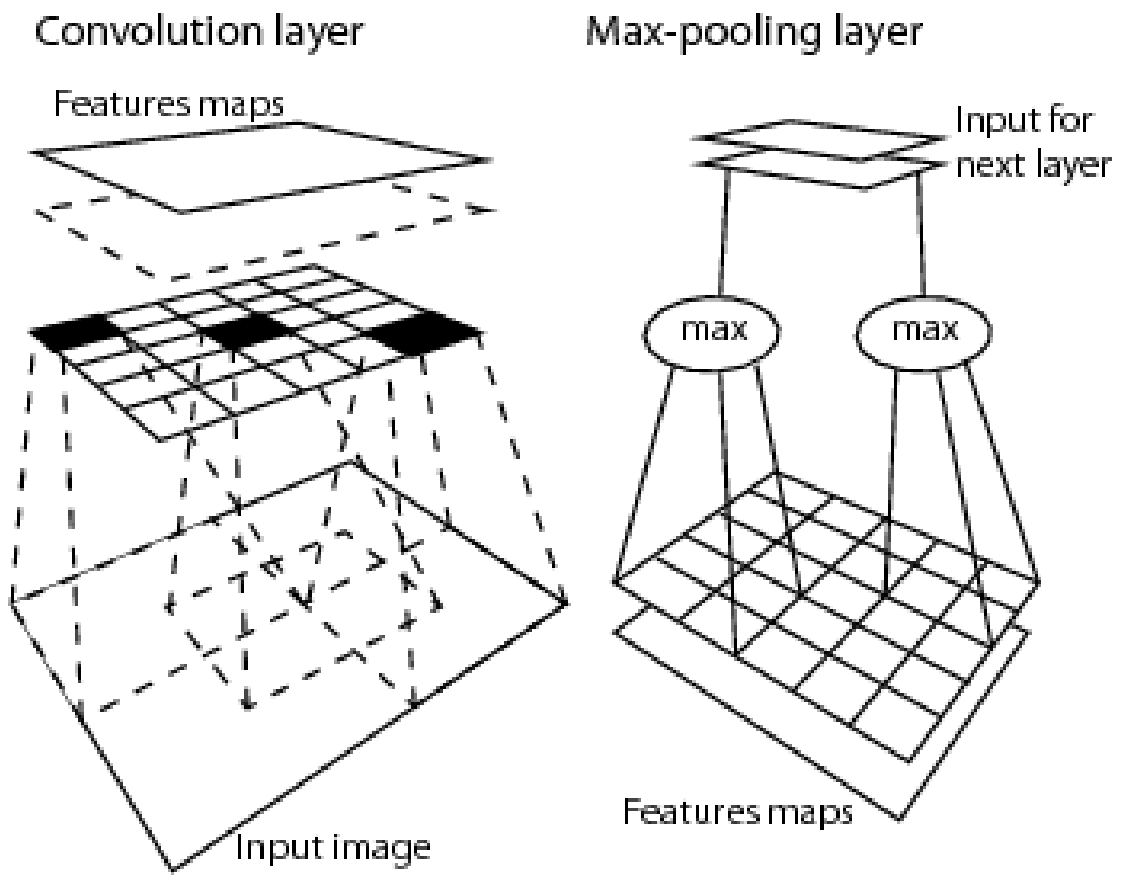

Figure 1-14. 2D image analysis as synonym of topology-preserving feature mapping in a (2D) image-domain. Activation domains of physically adjacent processing units in the 2D array of convolutional filters are spatially adjacent regions in the 2D visual field. Provided with a superior degree of biological plausibility in modelling 2D spatial topological and non-topological information, distributed processing systems capable of 2D image analysis, such as deep convolutional neural networks (DCNNs), typically outperform traditional 1D image analysis approaches. Will computers become as good as humans in 2D image analysis?

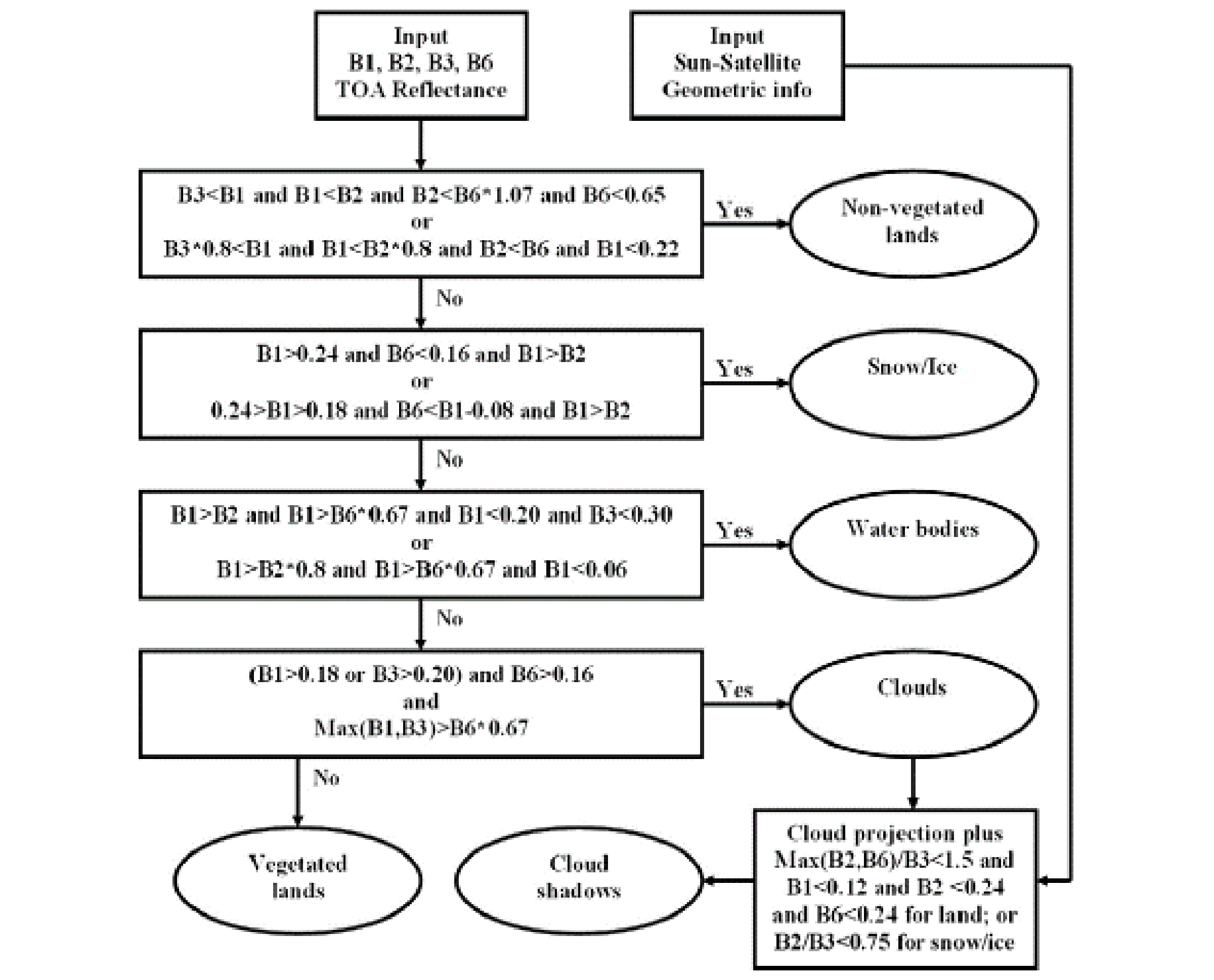

Figure 1-15. Courtesy of Luo et al. (2008). Canada Centre for Remote Sensing (CCRS)'s flow chart of a physical model-based per-pixel MODIS image classifier integrated into a clear-sky multi-temporal MODIS image compositing system in operating mode. Acronym B stands for MODIS Band.

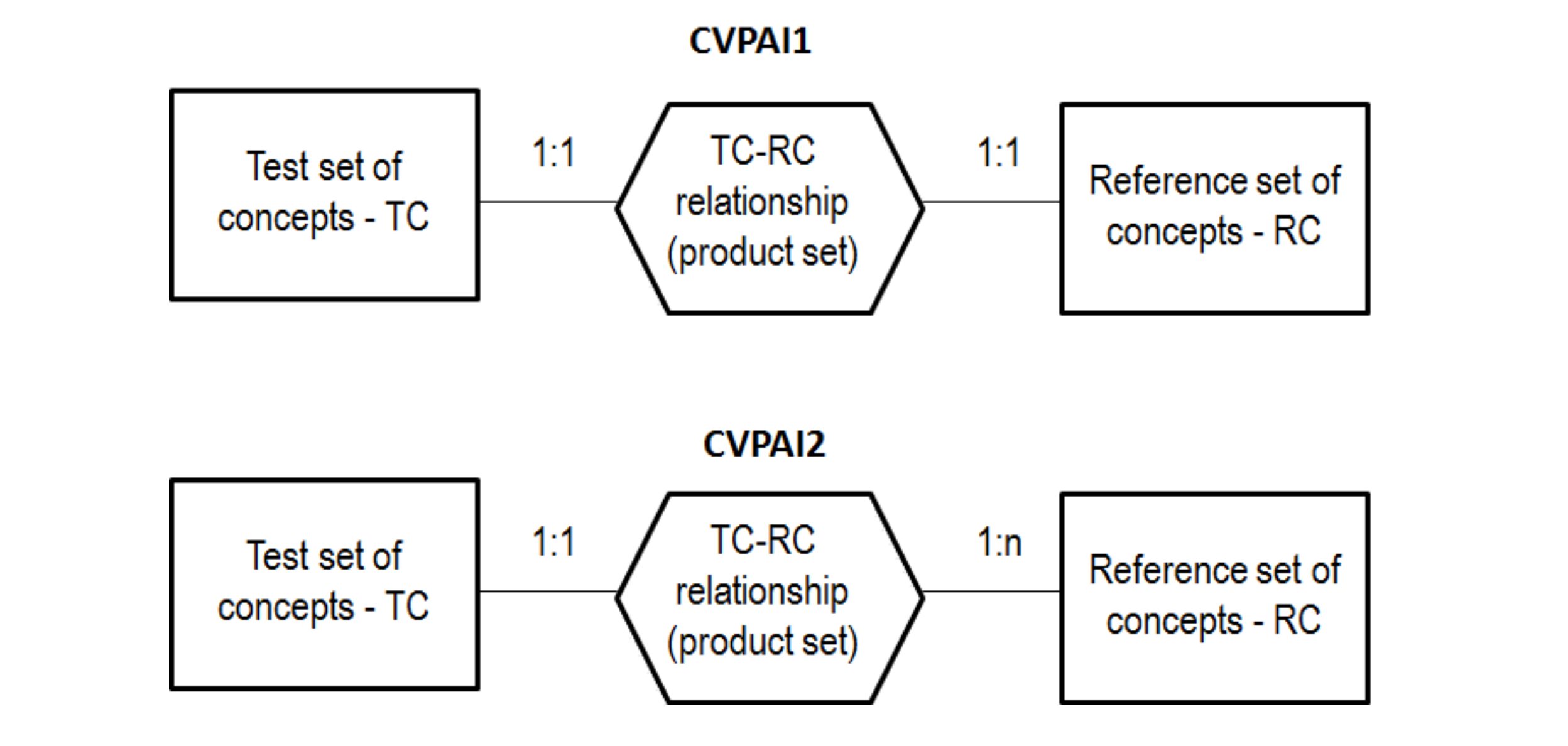

Figure 1-16. Entity-relationship conceptual model representation of the binary relationship R: A ⇒ B from set A = test categorical variable (TC) to set B = reference categorical variable (RC), provided with the min:max cardinality required to score maximum in range [0, 1] by the Categorical Variable-Pair Association Index (CVPAI) formulation 1 (CVPAI1) and 2 (CVPAI2) respectively, with CVPAI1 ≤ CVPAI2, i.e., the latter is a relaxed version of the former.

**Tables and table captions**

| SIAM, r88v6 | Input bands | Prior knowledge-based color map legends: Number of output spectral categories | | | |
|---|---|---|---|---|---|
| | | Fine discretization levels | Intermediate discretization levels | Coarse discretization levels | Inter-sensor discretization levels (*) |
| L-SIAM | 7 – B, G, R, NIR, MIR1, MIR2, TIR | 96 | 48 | 18 | 33<br>* employed for inter-sensor post-classification change/no-change detection |
| S-SIAM | 4 – G, R, NIR, MIR1 | 68 | 40 | 15 | |
| AV-SIAM | 4 – R, NIR, MIR1, TIR | 83 | 43 | 17 | |
| Q-SIAM | 4 – B, G, R, NIR | 61 | 28 | 12 | |

Table 1-1. The SIAM computer program is an EO system of systems scalable to any existing or future MS imaging sensors provided with radiometric calibration metadata parameters. It encompasses the following subsystems. (i) 7-band Landsat-like SIAM™ (L-SIAM™), with input channels Blue (B), Green (G), Red (R), Near Infra-Red (NIR), Medium IR1 (MIR1), Medium IR2 (MIR2), and Thermal IR (TIR). (ii) 4-band (channels G, R, NIR, MIR1) SPOT-like SIAM™ (S-SIAM™). (iii) 4-band (channels R, NIR, MIR1, and TIR) Advanced Very High Resolution Radiometer (AVHRR)-like SIAM™ (AV-SIAM™). (iv) 4-band (channels B, G, R, and NIR) QuickBird-like SIAM™ (Q-SIAM™).

| NLCD 2001/2006/2011 Classification Scheme (Legend), Level II | | | | LCCS-DP, level 1: A = Veg, B = Non-Veg, and level 2: 1 = Terrestrial, 2 = Aquatic |
|---|---|---|---|---|
| **Code** | **ID** | **Name** | **Land cover (LC) Class Definition** | **ID** |
| 11 | OW | Open water | OW: Areas of open water, generally with less than 25% cover of vegetation or soil | B4 - Non-vegetated aquatic |
| 12 | PIS | Perennial Ice/Snow | PIS: Areas characterized by a perennial cover of ice and/or snow, generally greater than 25% of total cover. | B4 |
| 21<br>22<br>23<br>24 | DOS<br>DLI<br>DMI<br>DHI | Developed, Open Space<br>Developed, Low Intensity<br>Developed, Medium Intensity<br>Developed, High Intensity | DOS: Includes areas with a mixture of some constructed materials, but mostly vegetation in the form of lawn grasses. Impervious surfaces account for less than 20 percent of total cover. These areas most commonly include large-lot single-family housing units, parks, golf courses, and vegetation planted in developed settings for recreation, erosion control, or aesthetic purposes.<br>DLI, DMI, DHI: refer to the "National Land Cover Database 2006 (NLCD2006)," Multi-Resolution Land Characteristics Consortium (MRLC), 2013. | B3 - Non-vegetated terrestrial / A1 - Vegetated terrestrial |
| 31 | BL | Barren Land (Rock/Sand/Clay) | BL: Barren areas of bedrock, desert pavement, scarps, talus, slides, volcanic material, glacial debris, sand dunes, strip mines, gravel pits and other accumulations of earthen material. Generally, vegetation accounts for less than 15% of total cover. As a consequence of this constraint, class BL covers only 1.21% of the CONUS total surface. | B3 |
| 41<br>42<br>43 | DF<br>EF<br>MF | Deciduous Forest<br>Evergreen Forest<br>Mixed Forest | DF: Areas dominated by trees generally greater than 5 meters tall, and greater than 20% of total vegetation cover. More than 75 percent of the tree species shed foliage simultaneously in response to seasonal change.<br>EF: Areas dominated by trees generally greater than 5 meters tall, and greater than 20% of total vegetation cover. More than 75 percent of the tree species maintain their leaves all year. Canopy is never without green foliage.<br>MF: Mixed Forest - Areas dominated by trees generally greater than 5 meters tall, and greater than 20% of total vegetation cover. Neither deciduous nor evergreen species are greater than 75 percent of total tree cover. | A1 |
| 51<br>52 | -<br>SS | Dwarf Scrub [2]<br>Scrub/Shrub | SS: Areas dominated by shrubs; less than 5 meters tall with shrub canopy typically greater than 20% of total vegetation. This class includes true shrubs, young trees in an early successional stage or trees stunted from environmental conditions. The aforementioned definition of class BL means that class SS may feature a vegetated cover which accounts for 15% of total cover or more. | A1/ B3 |
| 71<br>72<br>73<br>74 | GH<br>-<br>-<br>- | Grassland/Herbaceous<br>Sedge Herbaceous [2]<br>Lichens [2]<br>Moss [2] | GH: Areas dominated by grammanoid or herbaceous vegetation, generally greater than 80% of total vegetation. These areas are not subject to intensive management such as tilling, but can be utilized for grazing. The aforementioned definition of class BL means | A1/B3 |
| | | | that class GH may feature a vegetated cover which accounts for 15% of total cover or more. | |
| 81<br>82 | PH<br>CC | Pasture/Hay<br>Cultivated Crops | PH: Areas of grasses, legumes, or grass-legume mixtures planted for livestock grazing or the production of seed or hay crops, typically on a perennial cycle. Pasture/hay vegetation accounts for greater than 20 percent of total vegetation.<br>CC: Areas used for the production of annual crops, such as corn, soybeans, vegetables, tobacco, and cotton, and also perennial woody crops such as orchards and vineyards. Crop vegetation accounts for greater than 20% of total vegetation. This class also includes all land being actively tilled. | A1 |
| 90<br>95 | WW<br>EHW | Woody Wetlands<br>Emergent Herbaceous Wetland | WW: Areas where forest or shrubland vegetation accounts for greater than 20 percent of vegetative cover and the soil or substrate is periodically saturated with or covered with water.<br>EHW: Areas where perennial herbaceous vegetation accounts for greater than 80% of vegetative cover and the soil or substrate is periodically saturated with or covered with water. | A2 – Vegetated aquatic |

Table 1-2. Definition of the NLCD 2001/2006/2011 classification taxonomy, Level II. [2]Alaska only. For further details, refer to the "National Land Cover Database 2006 (NLCD2006)," Multi-Resolution Land Characteristics Consortium (MRLC), 2013.. The right column instantiates a possible binary relationship R: $A \Rightarrow B \subseteq A \times B$ from set A = NLCD legend to set B = 2-level 4-class Dichotomous Phase (DP) taxonomy of the Food and Agriculture Organization of the United Nations (FAO) - Land Cover Classification System (LCCS) (Di Gregorio and Jansen 2000), refer to Figure 1-2.

| | | | Target classes of individuals (entities in a conceptual model for knowledge representation built upon an ontology language) | | |
|---|---|---|---|---|---|
| | | | Class 1, Water body | Class 2, Tulip flower | Class 3, Italian tile roof |
| Color names | black | | | √ | |
| | blue | | √ | √ | |
| | brown | | √ | √ | √ |
| | grey | | | | |
| | green | | √ | √ | |
| | orange | | | √ | |
| | pink | | | √ | |
| | purple | | | √ | |
| | red | | | √ | √ |
| | white | | | √ | |
| | yellow | | | √ | |

Table 1-3. Example of a binary relationship $R: A \Rightarrow B \subseteq A \times B$ from set A = DictionaryOfColorNames, with cardinality $|A| = a$ = ColorDictionaryCardinality = 11, and the set B = LegendOfObjectClassNames, with cardinality $|B| = b$ = ObjectClassLegendCardinality = 3. The latter dictionary is a superset of the typical taxonomy of land cover (LC) classes adopted by the RS community. "Correct" entry-pairs (marked with √) must be: (i) selected by domain experts based on a hybrid combination of deductive prior beliefs with inductive evidence from data (refer to Table 1-5) and (ii) community-agreed upon.

| **Index** | **Spectral Categories** | **Spectral Rule (**based on reflectance measured at Landsat TM central wave bands: b1 is located at 0.48 μm, b2 at 0.56 μm, b3 at 0.66 μm, b4 at 0.83 μm, b5 at 1.6 μm, b7 at 2.2 μm) | **Pseudo-color** |
|---|---|---|---|
| 1 | Snow/ice | $b4/b3 \leq 1.3$ AND $b3 \geq 0.2$ AND $b5 \leq 0.12$ | |
| 2 | Cloud | $b4 \geq 0.25$ AND $0.85 \leq b1/b4 \leq 1.15$ AND $b4/b5 \geq 0.9$ AND $b5 \geq 0.2$ | |
| 3 | Bright bare soil / sand / cloud | $b4 \geq 0.15$ AND $1.3 \leq b4/b3 \leq 3.0$ | |
| 4 | Dark bare soil | $b4 \geq 0.15$ AND $1.3 \leq b4/b3 \leq 3.0$ AND $b2 \leq 0.10$ | |
| 5 | Average vegetation | $b4/b3 \geq 3.0$ AND ($b2/b3 \geq 0.8$ OR $b3 \leq 0.15$) AND $0.28 \leq b4 \leq 0.45$ | |
| 6 | Bright vegetation | $b4/b3 \geq 3.0$ AND ($b2/b3 \geq 0.8$ OR $b3 \leq 0.15$) AND $b4 \geq 0.45$ | |
| 7 | Dark vegetation | $b4/b3 \geq 3.0$ AND ($b2/b3 \geq 0.8$ OR $b3 \leq 0.15$) AND $b3 \leq 0.08$ AND $b4 \leq 0.28$ | |
| 8 | Yellow vegetation | $b4/b3 \geq 2.0$ AND $b2 \geq$ _b3 AND $b3 \geq 8.0$ AND $b4/b5 \geq 1.5$ [a] | |
| 9 | Mix of vegetation / soil | $2.0 \leq b4/b3 \leq 3.0$ AND $0.05 \leq b3 \leq 0.15$ AND $b4 \geq 0.15$ | |
| 10 | Asphalt / dark sand | $b4/b3 \leq 1.6$ AND $0.05 \leq b3 \leq 0.20$ AND $0.05 \leq b4 \leq 0.20$[a] AND $0.05 \leq b5 \leq 0.25$ AND $b5/b4 \geq 0.7$[a] | |
| 11 | Sand / bare soil / cloud | $b4/b3 \leq 2.0$ AND $b4 \geq 0.15$ AND $b5 \geq 0.15$[a] | |
| 12 | Bright sand / bare soil / cloud | $b4/b3 \leq 2.0$ AND $b4 \geq 0.15$ AND ($b4 \geq 0.25$b OR $b5 \geq 0.30$[b]) | |
| 13 | Dry vegetation / soil | ($1.7 \leq b4/b3 \leq 2.0$ AND $b4 \geq 0.25$[c]) OR ($1.4 \leq b4/b3 \leq 2.0$ AND $b7/b5 \leq 0.83$[c]) | |
| 14 | Sparse veg. / soil | ($1.4 \leq b4/b3 \leq 1.7$ AND $b4 \geq 0.25$[c]) OR ($1.4 \leq b4/b3 \leq 2.0$ AND $b7/b5 \leq 0.83$ AND $b5/b4 \geq 1.2$[c]) | |
| 15 | Turbid water | $b4 \leq 0.11$ AND $b5 \leq 0.05$[a] | |
| 16 | Clear water | $b4 \leq 0.02$ AND $b5 \leq 0.02$[a] | |
| 17 | Clear water over sand | $b3 \geq 0.02$ AND $b3 \geq b4 + 0.005$ AND $b5 \leq 0.02$[a] | |
| 18 | Shadow | | |
| 19 | Not classified (outliers) | | |

| |
|---|
| [a] These expressions are optional and only used if b5 is present. [b] Decision rule depends on presence of b5. [c] Decision rule depends on presence of b7. |
| Table 1-4. Rule set (structural knowledge) and order of presentation of the rule set (procedural knowledge) adopted by the prior knowledge-based MS reflectance space quantizer called Spectral Classification of surface reflectance signatures (SPECL), implemented within the ATCOR commercial software toolbox (Dorigo et al. 2009; Richter and D. Schläpfer 2012a, 2012b). |

**STEP 1. Dictionary-pair relationship, multivariate occurrence distributions**

| | | *Reference Classification (RC)* | | | | |
|---|---|---|---|---|---|---|
| | | EvergreenF | DeciduousF | Others | | |
| ***Test Classification (TC)*** | Vegetation | 10 | 30 | 60 | 100 | |
| | Cloud | 2 | 0 | 10 | 12 | |
| | Unknowns | 0 | 5 | 100 | 105 | Tot. |
| | | 12 | 35 | 170 | | 217 |

**STEP 2. Dictionary-pair relationship, multivariate probability distributions**

| | | *Reference Classification (RC)* | | | | |
|---|---|---|---|---|---|---|
| | | EvergreenF | DeciduousF | Others | | |
| ***Test Classification (TC)*** | Vegetation | 0.046082949 | 0.138248848 | 0.276498 | 0.460829 | |
| | Cloud | 0.00921659 | 0 | 0.046083 | 0.0553 | |
| | Unknowns | 0 | 0.023041475 | 0.460829 | 0.483871 | Tot. |
| | | 0.055299539 | 0.161290323 | 0.78341 | | 1 |

**STEP 3. Dictionary-pair relationship, cond. prob. (RC|TC)**

| | | *Reference Classification (RC)* | | | |
|---|---|---|---|---|---|
| | | EvergreenF | DeciduousF | Others | |
| ***Test Classification (TC)*** | Vegetation | 0.1 | 0.3 | 0.6 | 1 |
| | Cloud | 0.166666667 | 0 | 0.833333 | 1 |
| | Unknowns | 0 | 0.047619048 | 0.952381 | 1 |

**STEP 4. Crisp membership function(RC|TC) > TH1 = 0.09.**

| | | *Reference Classification (RC)* | | |
|---|---|---|---|---|
| | | EvergreenF | DeciduousF | Others |
| ***Test Classification (TC)*** | Vegetation | 1 | 1 | 1 |
| | Cloud | 1 | 0 | 1 |
| | Unknowns | 0 | 0 | 1 |

**STEP 5. Dictionary-pair relationship, cond. prob. (TC|RC)**

| | | *Reference Classification (RC)* | | |
|---|---|---|---|---|
| | | EvergreenF | DeciduousF | Others |
| ***Test Classification (TC)*** | Vegetation | 0.833333333 | 0.857142857 | 0.352941 |
| | Cloud | 0.166666667 | 0 | 0.058824 |
| | Unknowns | 0 | 0.142857143 | 0.588235 |
| | | 1 | 1 | 1 |

**STEP 6. Crisp membership function(TC|RC) > TH2 = 0.06 <= TH1 = 0.09.**

| | | *Reference Classification (RC)* | | |
|---|---|---|---|---|
| | | EvergreenF | DeciduousF | Others |
| ***Test Classification (TC)*** | Vegetation | 1 | 1 | 1 |
| | Cloud | 1 | 0 | 0 |
| | Unknowns | 0 | 1 | 1 |

**STEP 7. OR{Crisp membership function(TC|RC), Crisp membership function(RC|TC)}**

| | | *Reference Classification (RC)* | | |
|---|---|---|---|---|
| | | EvergreenF | DeciduousF | Others |
| ***Test Classification (TC)*** | Vegetation | 1 | 1 | 1 |
| | Cloud | 1 | 0 | 1 |
| | Unknowns | 0 | 1 | 1 |

**STEP 8. Top-down (driven-by-prior knowledge) scrutiny of bottom-up (data-driven) "temporary correct" or "temporary non-correct" cells**

| | | *Reference Classification (RC)* | | |
|---|---|---|---|---|
| | | EvergreenF | DeciduousF | Others |
| ***Test Classification (TC)*** | Vegetation | 1 | 1 | 1 |
| | Cloud | 0 | 0 | 1 |
| | Unknowns | 0 | 0 | 1 |

Table 1-5. 8-step guideline for best practice in the identification of a dictionary-pair relationship based on a hybrid combination of prior beliefs, if any, with frequentist inference.